\documentclass[10pt,journal,compsoc]{IEEEtran}
\ifCLASSOPTIONcompsoc
  \usepackage[nocompress]{cite}
\else
  \usepackage{cite}
\fi
\ifCLASSINFOpdf
\else
\fi
\usepackage[T1]{fontenc}% optional T1 font encoding
\usepackage{epsfig}
\usepackage{amsmath,bm}
\usepackage{algorithm}
\usepackage{algorithmic}
\usepackage{subfigure}
\usepackage{epstopdf}
\usepackage{enumerate}
\usepackage{tabularx}
\usepackage{amsthm}
\usepackage{booktabs}
\usepackage{balance}
\usepackage{xspace}
\usepackage{graphicx}
\usepackage{amsfonts}
\usepackage{balance}
\usepackage{color}

\theoremstyle{definition}
\newtheorem{definition}{Definition}[section]

\newcommand{\tool}{\texttt{SL-GAD}\xspace}
\newcommand{\hl}{\textcolor{black}}

\begin{document}
\title{Generative and Contrastive Self-Supervised Learning for Graph Anomaly Detection}

\author{Yu Zheng, Ming Jin, 
Yixin Liu, Lianhua Chi, Khoa T. Phan, Yi-Ping Phoebe Chen% <-this % stops a space
\IEEEcompsocitemizethanks{
\IEEEcompsocthanksitem Y. Zheng, L. Chi, K. T. Phan, and Y-P. P. Chen are with Department of Computer Science and Information Technology, La Trobe University, Melbourne Australia \protect\\
E-mail: Yu.Zheng@latrobe.edu.au, l.chi@latrobe.edu.au, K.Phan@latrobe.edu.au, phoebe.chen@latrobe.edu.au
\IEEEcompsocthanksitem M. Jin and Y. Liu  are with the Department of Data Science and AI, Faculty of IT, Monash University, Clayton, VIC 3800, Australia\protect\\
E-mail:  ming.jin@monash.edu; yixin.liu@monash.edu;
\IEEEcompsocthanksitem Corresponding Authors: Ming Jin and Lianhua Chi.
}% <-this % stops an unwanted space
}


\markboth{Journal of \LaTeX\ Class Files,~Vol.~14, No.~8, August~2015}%
{Shell \MakeLowercase{\textit{et al.}}: Bare Demo of IEEEtran.cls for Computer Society Journals}



\IEEEtitleabstractindextext{%
\begin{abstract}
Anomaly detection from graph data has drawn much attention due to its practical significance in many critical applications including cybersecurity, finance, and social networks. Existing data mining and machine learning methods are either shallow methods that could not effectively capture the complex interdependency of graph data or graph autoencoder methods that could not fully exploit the contextual information as supervision signals for effective anomaly detection. To overcome these challenges, in this paper, we propose a novel method, Self-Supervised Learning for Graph Anomaly Detection (\tool). Our method constructs different contextual subgraphs (views) based on a target node and employs two modules, \textit{generative attribute regression} and \textit{multi-view contrastive learning} for anomaly detection.  While the 
 \textit{generative attribute regression}
module allows us to capture the anomalies in the attribute space, the \textit{multi-view contrastive learning} module can exploit richer structure information from multiple subgraphs, thus abling to capture the anomalies in the structure space, mixing of structure, and attribute information. We conduct extensive experiments on six benchmark datasets and the results demonstrate that our method outperforms  state-of-the-art methods by a large margin.
\end{abstract}

}

\maketitle

\IEEEdisplaynontitleabstractindextext

% \fi
\IEEEpeerreviewmaketitle

\IEEEraisesectionheading{\section{Introduction}\label{sec:introduction}}

\IEEEPARstart{R}{ecent} years have witnessed increasing domains that continuously generate complex, interdependent,  and connected data, represented in the form of graphs or networks. Typical examples include social networks, biological networks, traffic networks, and financial transaction networks, to name a few. Data mining and analysis from these graph-structured data have drawn much attention, particularly for the task of graph anomaly detection, where the goal is to identify patterns (e.g., nodes, edges, subgraphs)  which differ significantly from the majority patterns in the graphs. For instance, in the financial transaction network, it is critically important to identify the abnormal edges (fraudulent transactions) between two accounts \cite{pourhabibi2020fraud}. In a social network, it is also crucial to detect the abnormal nodes  (social bots) as they may spread rumours over social networks \cite{latah2020detection}.

\hl{Detecting graph anomalies however is a challenging task because many graphs contain complex linkage (structure) information as well as node attribute information.} As a result, anomalies can be hidden in the structure space, attribute space, and the mix of both. Furthermore, in many cases, the ground truths of the anomalies are unknown, \hl{rendering} many supervised classification approaches not applicable.  These two challenges have motivated increasing efforts \hl{for efficient anomaly detection in recent years}, ranging from shallow methods to deep representation methods for anomaly detection, in a purely unsupervised manner.

The shallow methods mainly focus on defining anomaly quantify measures for graphs and \hl{developing methods} to capture the anomaly based on these measures. Perozzi and Akoglu \cite{amen_perozzi2016scalable} \hl{propose a normality measure} to evaluate neighbourhoods both internally and externally by considering both attributes and graph structure\hl{, where Anomaly Mining of Entity Neighborhoods (AMEN) is proposed} to optimise the measure to get the anomaly score.  Noticing that the residual of regression plays an important role for qualifying the anomaly score, Li \textit{et al.} \cite{radar_li2017radar} propose a Radar framework that learns a linear regression function to fit the node attributes regularized by the network structure. The residual from the regression function is used as a score to measure the anomaly.  Similarly, Peng \textit{et al.} \cite{anomalous_peng2018anomalous} propose a joint modelling approach to conduct attribute selection and anomaly detection \hl{using the residual.} While being simple, \hl{these shallows are not able to model (or capture)} the complex interdependent relations of graphs.

Deep learning-based approaches have shown impressive progress in many domains, including image, text, and graphs. For the task of graph anomaly detection, autoencoder becomes a popular choice as it is a purely unsupervised framework and fits settings where no ground-truth label is available. Specifically, Dominant \cite{dominant_ding2019deep} employs a graph convolution network (GCN) to encode both structure and node content into a latent embedding, based on which both attribute and structure reconstruction decoders are used. The anomaly score is calculated by the weighted sum of the
reconstruction errors of attribute and structure. SpaceAE \cite{specae_li2019specae} employs a spectral autoencoder with  a density estimation model for anomaly detection. AEGIS \cite{ding2020inductive} further generalises the graph autoencoder to the inductive setting where unseen anomalies may exist.

However, existing graph autoencoder based methods \hl{do} not fully exploit the \textit{contextual information} (e.g., neighbouring nodes or subgraphs) which \hl{is} critically important for anomaly detection. These autoencoder approaches typically aim to reconstruct the whole graph structure or attributes for every single node.  As \hl{anomaly} detection aims to ``identify patterns in data that do not conform to \textit{expected behaviour}” \cite{chandola2009anomaly}, it is natural to define the expected behaviour based the contextual information, which is largely ignored in existing methods. Furthermore, these algorithms \hl{do} not fully exploit the available information as supervision signals to learn their models. The learning objective in these methods is mostly focused on the node level (e.g., learning node-level embedding directly from graph autoencoder). The subgraph information of a target node can provide more supervision signals and that are also not well exploited for graph anomaly detection. 

% As there is rich structure information as well as attribute information in the network, how to effectively utilise the available information in networks has been a long-standing 

% suffer two limitations: 
% \begin{itemize}
%     \item \textit{Contextual information}. Existing methods did not fully exploit the \textit{contextual information} (e.g., neighbouring nodes or subgraphs) which are critically important for anomaly detection. These autoencoder approaches typically aim to reconstruct the whole graph structure or attributes for each single node.  As Anomaly detection aims to ``identify patterns in data that do not conform to \textit{expected behaviour}” \cite{chandola2009anomaly}, it is naturally to define the expected behaviour based the the contextual information, which is unfortunately largely ignored in existing methods. 
%     \item \textit{Self-supervision signals}. Existing methods did not fully exploit the available information as a supervision signals to learn their models. As the most intuitive way
% \end{itemize}

\begin{figure}[t]
\centering
\includegraphics[width=0.45\textwidth]{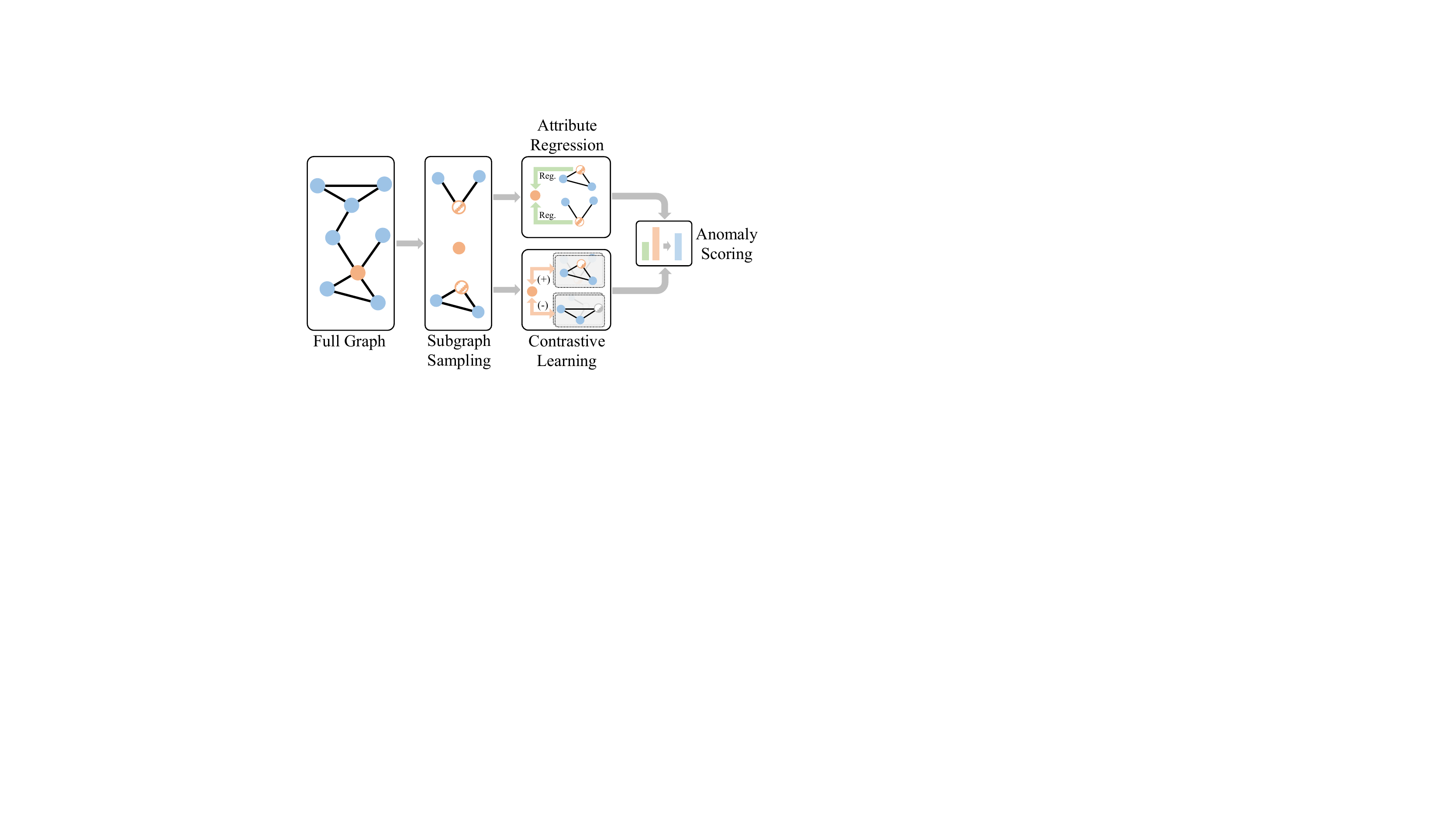}
\caption{The workflow chart of $\tool$ on self-supervised graph anomaly detection. The abnormality of a node is estimated on twofold perspectives, namely generative and contrastive graph anomaly scoring.}
\label{fig:workflow}
\end{figure}

Based on these observations, in this paper, we present a novel self-supervised algorithm, \tool, for graph anomaly detection. \hl{Our theme is to construct different surrounding contexts (subgraphs)} of a target node and employ self-supervised learning strategies to make comparison and obtain the anomaly score for each node. Specifically, we first sample different views (local subgraphs) centralised on each target node as the contextual information. These \hl{views} are then fed into a graph neural network (GNN) encoder to learn the latent representation of each node. After this, we employ two modules, namely \textit{generative attribute reconstruction} %\textit{attribute regression}%
and \textit{multi-view contrastive learning}, to fully exploit the available information in a self-supervised manner. By reconstructing node attributes with a GNN decoder, the 
\textit{generative attribute reconstruction} %\textit{attribute regression}% 
module is able to capture anomalies in the attribute space. 
%By comparing different subgraphs based on the target node, the \textit{multi-view contrastive learning} module can identify the differences between the target node and its contextual subgraphs, thus capture anomalies in the structure space and mix of structure and content information. 
\hl{By directly comparing a target node with its surrounding contexts, the \textit{multi-view contrastive learning} module can capture anomalies in the structure space and mix of both structure and content information.}
Finally, the anomaly score is calculated based on both 
generative and contrastive %attribute regression and contrastive learning% 
modules to provide a comprehensive score to qualify the \hl{abnormality} of each node. Experimental results on six datasets show the effectiveness of our algorithm. A workflow chart of our model is given in Figure \ref{fig:workflow}.

The main contribution of this paper can be summarised as follows.
\begin{itemize}
    \item We propose a novel self-supervised based method for graph data. By developing a subgraph-based contrastive learning module, our method can effectively exploit the useful information in graphs.
    
    \item We develop a new method for graph anomaly detection. By developing both generative attribute regression and multi-view contrastive learning   mechanisms, our method provides a new way to qualify the anomaly score for each node for anomaly detection.
    
    \item We have conducted extensive experimental results on six datasets to test our design for graph anomaly detection. Experimental results show that our algorithm outperforms \hl{state-of-the-art algorithms} by a large margin.
\end{itemize}

The rest of the paper is organised as follows: Section 2 \hl{reviews} the related work. Section 3 depicts the  definition of the problem. Section 4 presents the proposed method. Section 5 discusses the experimental results and we conclude this paper in Section 6. 


\section{Related Work}
This work is closely related to graph anomaly detection, self-supervised learning, and graph representation learning. We briefly review these related works in this section.
\subsection{Anomaly detection}
Anomaly detection has been a long-standing research topic \cite{chandola2009anomaly, pang2020deep}. Aiming to identify patterns that significantly differ from the expected patterns, research in this area has been evolving from traditional statistic methods such as local outlier factor \cite{breunig2000lof} and one-class support vector machine  \cite{scholkopf2001estimating} to deep learning approaches such as Outlier Exposure (OE) \cite{hendrycks2018deep} and Deep Semi-Supervised Anomaly Detection (Deep SAD) \cite{ruff2019deep}. While these methods typically target data in the Euclidean domain, recently, anomaly detection from graph structure data which are outside Euclidean space draws increasing attention \cite{amen_perozzi2016scalable, radar_li2017radar, anomalous_peng2018anomalous, liuanomaly}. Because how to measure the anomaly is an important problem for anomaly detection in graphs, Perozzi and Akoglu \cite{amen_perozzi2016scalable} defined the normality as a measure and proposed an AMEN approach for anomaly detection. Exploiting the residual for anomaly detection, Li \textit{et al.} proposed the Radar approach \cite{radar_li2017radar} and Peng \textit{et al.} proposed the  Anomalous approach \cite{anomalous_peng2018anomalous} which employ matrix regression approach and regard nodes with large residual as anomalies. Recently, deep approaches have been also applied in graphs for anomaly detection \cite{dominant_ding2019deep,specae_li2019specae,ding2020inductive}.  These methods typically employ a graph autoencoder to embed the nodes in a latent space, and then reconstruct the graph information. The reconstruction errors are used to detect anomalies. Recently, Liu \textit{et al.} proposed a self-supervised approach called CoLA \cite{liuanomaly}, which exploits the local information from network data by sampling pairs of instance, and employs the contrastive learning to learn the node representation. The abnormal score is calculated based on the predicted scores on the contrastive pairs. However, CoLA only utilizes self-supervised learning in a contrastive manner to capture the anomaly patterns, which \hl{limits} supervision signals to learn the model due to the lack of generative self-supervised learning.
%CoLA has achieved state-of-the-art performance in graph benchmark datasets.   

\subsection{Self-supervised learning}
Self-supervised learning (SSL) \cite{cl_survey_liu2020selfsupervised} is a new learning paradigm which aims to learn a neural model from the unsupervised data itself without using human-annotated labels. By training on well-designed \textit{pretext
tasks}, SSL enables a model to learn better representations that can be generalized to downstream tasks. SSL has achieved great success on computer vision (CV) \cite{jing2020self}
and natural language processing (NLP)  \cite{cl_survey_liu2020selfsupervised}. Recently, SSL has been extended to graph domains (please refer to \cite{liu2021graph} for a comprehensive survey for graph SSL). 
In particular, Veličković \textit{et al.}  proposed the first contrastive learning algorithm, Deep Graph Infomax (DGI) \cite{velickovic2019deep}, to learn the embedding form graph data in an unsupervised manner. 
Hassani and Khasahmadi proposed an approach MVGRL \cite{hassani2020contrastive} which performs contrastive learning on graphs from first-order neighbours and a graph diffusion on two views.
Wan \textit{et al.} proposed an algorithm CG$^3$ \hl{\cite{wan2021contrastive}} which exploits contrastive learning based on both graph structure and limited labelled information among \hl{nodes. 
By capitalizing on the idea of self-knowledge distillation, Jin \textit{et al.} proposed a method MERIT \cite{Jin2021MultiScaleCS} to enrich supervision signals by maximizing the agreement of node embeddings across different views and networks.
% SUBG-CON \cite{Jiao2020SubgraphCF}, on the other hand, proposed a scalable graph SSL algorithm by composing of the top-$k$ neighbor sampling-based graph augmentations and the objective of MVGRL, achieving the state-of-the-art performance on node classification tasks.
JOAO \cite{You2021JOAO}, on the other hand, proposed to automatically learn graph augmentations with the self-supervised model, which alleviates the reliance on the design of augmentations.
However, while} all these algorithms only focus on learning the representation for nodes in graphs, they have not been applied for anomaly detection.

\begin{table}[t]
	\small
	\centering
	\caption{Summary of the primary notations.} 
	\begin{tabular}{ p{65 pt}<{\centering} | p{165 pt}}  
		\toprule[1.0pt]
		Symbols & Description  \\
		\cmidrule{1-2}
		$\mathcal{G}=(\mathbf{X}, \mathbf{A})$ & An attributed graph \\
		$\mathcal{V}, \mathcal{E}$ & The node and edge set of $\mathcal{G}$ \\
		$\mathbf{A} \in \mathbb{R}^{N \times N}$ & The adjacency matrix of $\mathcal{G}$ \\
		$\mathbf{X} \in \mathbb{R}^{N \times D}$ & The node features matrix of $\mathcal{G}$ \\
		$\mathbf{x}_i \in \mathbb{R}^{D}$ & \hl{The feature vector of $v_i$ that $ \mathbf{x}_i \in \mathbf{X}$} \\
		$\mathcal{N}(v_i)$ & \hl{The neighbors of node $v_i \in \mathcal{V}$} \\
		\cmidrule{1-2}
		$v_t \in \mathcal{V}$ & A selected target node \\
		$\mathcal{G}_{t}^{\phi_1}$, $\mathcal{G}_{t}^{\phi_2}$ & Two generated graph views of $v_t$ \\
		\hl{$\mathbf{X}_{t}^{\phi_i} \in \mathbb{R}^{K \times D}$} & The node feature matrix of $\mathcal{G}_{t}^{\phi_i}$ \\
		\hl{$\widehat{\mathbf{X}}_{t}^{\phi_i}[-1,:] \in \mathbb{R}^{D}$} & The reconstructed feature vector of anonymized target node in $\mathcal{G}_{t}^{\phi_i}$ \\
		\hl{$\mathbf{A}_{t}^{\phi_i} \in \mathbb{R}^{K \times K}$} & The adjacency matrix of $\mathcal{G}_{t}^{\phi_i}$ \\
		$f(v_t)$ & The anomaly score of $v_t$ \\
		\cmidrule{1-2}
		$\mathbf{h}_{t} \in \mathbb{R}^{D'}$ & The embedding vector of target node $v_t$ \\
		$\mathbf{g}_{\phi_i} \in \mathbb{R}^{D'}$ & The graph embedding vector of $\mathcal{G}_t^{\phi_i}$ \\
		$\mathbf{H}_{\phi_i} \in \mathbb{R}^{K \times D'}$ & The node embedding matrix of $\mathcal{G}_t^{\phi_i}$ \\
		$\mathbf{H}_{\phi_i}[j,:] \in \mathbb{R}^{D'}$ & The embedding vector of $v_j$ in $\mathbf{H}_{\phi_i}$ \\
		$\mathbf{H}_{\phi_i}^{(\mathit{l})} \in \mathbb{R}^{K \times D'_{\mathit{l}}}$ & The node embedding matrix of $\mathcal{G}_t^{\phi_i}$ on the $l$-th GNN layer \\
		$\mathbf{H}_{\phi_i}^{(\mathit{l})}[j,:] \in \mathbb{R}^{D'_{\mathit{l}}}$ & The embedding vector of $v_j$ in $\mathbf{H}_{\phi_i}^{(\mathit{l})}$ \\
		$\mathbf{W}_{enc} \in \mathbb{R}^{D \times D}$ & The \hl{trainable} parameter matrix of graph encoder \\
		$\mathbf{W}_{dec} \in \mathbb{R}^{D \times D}$ & The \hl{trainable} parameter matrix of graph decoder \\	
		$\mathbf{W}_{s} \in \mathbb{R}^{D' \times D'}$ & The \hl{trainable} parameter matrix of contrastive discriminator \\
		\cmidrule{1-2}
		$N$ & The number of nodes in $\mathcal{G}$ \\
		$K$ & The number of nodes in graph views \\
		$D$ & The dimension of node features in $\mathcal{G}$ \\
 		$D'$ & The dimension of embeddings in $\mathbf{H}_{\phi_i}$  \\
		$D'_{\mathit{l}}$ & The dimension of embeddings in $\mathbf{H}_{\phi_i}^{(\mathit{l})}$ \\
		$R$ & The number of evaluation rounds to calculate final anomaly scores \\
		\bottomrule[1.0pt]
	\end{tabular}
	\label{table:notation}
\end{table}

\subsection{Graph representation learning}
Our work is also related to \textit{graph representation learning}, where the goal is to learn a representation for each node or the entire graph, so that downstream graph analysis tasks (such as node classification and anomaly detection) can be easily performed. Graph neural networks (GNNs) \cite{gnn_survey_wu2020comprehensive}, particularly graph convolutional networks \cite{gcn_kipf2017semi,gat_ve2018graph}, have achieved great success for this task. Kipf and Welling proposed the graph convolutional networks (GCNs) \cite{gcn_kipf2017semi} which employs a two-level architecture and performs message passing in the \hl{spectral} domain, showing impressive performance in the node classification task. Veličković \textit{et al.} proposed a graph attention network (GAT)  \cite{gat_ve2018graph} which employs a neural network to automatically learn the weights (attentional scores) of each neighbour in the process of message passing, further improving the performance of GCN.
To improve the scalability of graph neural networks, Hamiltion \textit{et al.} proposed a GraphSage algorithm \cite{sage_hamilton2017inductive} which performs sampling for the neighbour message aggregation. \hl{Differently, Frasca \textit{et al.} proposed a method SIGN based on graph convolutional filters with different size \cite{sign_icml_grl2020}, which alleviates the reliance on graph sampling.}
To improve the robustness of graph representation learning, Pan \textit{et al.} proposed an adversarial regularised graph autoencoder (ARGA) which employs an adversarial training method to regularise the embedding in the latent space. \hl{Geisler \textit{et al.}, on the other hand, proposed a robust aggregation function to learn robust graph representations \cite{geisler2020reliable}.}
To improve the rigidness and inflexibility of deterministic classification functions employed in existing GNN methods, Wang \textit{et al.} proposed a novel framework named Graph Stochastic Neural Networks (GSNN) \cite{wang2020graph}, which aims to model the uncertainty of the classification function by simultaneously learning a family of functions, i.e., a stochastic function. 
As many GNNs lack the flexibility to model intrinsic complex graph geometry \hl{by embedding graphs into either Euclidean or hyperbolic spaces}, a graph geometry interaction learning algorithm (GIL) \cite{zhu2020graph} is proposed recently to utilize the strength of both Euclidean and hyperbolic geometries. Wu \textit{et al.} proposed an algorithm \cite{wu2021learning} to handle data in a positive and unlabelled learning setting, in which only part of the nodes are labelled as positive nodes and the majority of nodes are unlabelled nodes.  Graph representation learning techniques have also been widely applied in heterogeneous networks \cite{zhu2019relation}, spatial-temporal networks \cite{wu2020connecting}, \hl{community detection \cite{jin2021survey}}, and image classification \cite{wan2020hyperspectral}. However, existing GNNs approaches are mostly focused on generic graph representation learning. Employing GNNs for anomaly detection is still under-explored. By integrating GNNs with self-supervised learning, we will develop a new approach for graph anomaly detection in this paper. 

\section{Problem Definition}
In this section, \hl{we introduce the unsupervised graph anomaly detection problem} and notations used in the paper. Specifically, we use bold uppercase and lowercase letters to denote matrices and vectors, respectively. All important notations have been summarized in Table \ref{table:notation}. \hl{Graph and graph neural network (GNN) are defined as follows:}

\begin{figure*}[t]
\centering
\includegraphics[width=0.97\textwidth]{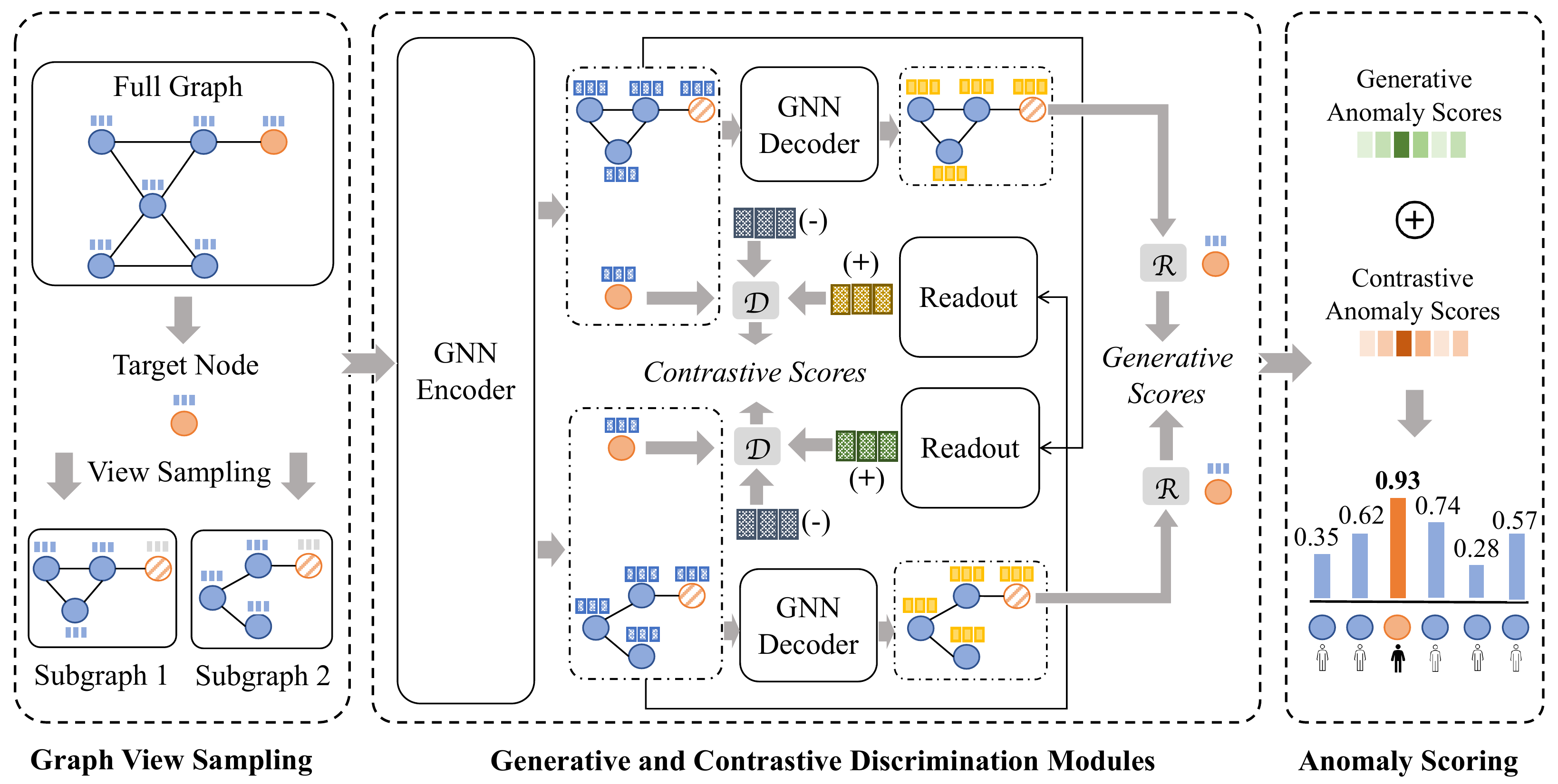}
\caption{The conceptual framework of $\tool$, which consists of three main components: Graph view sampling, self-supervised discrimination, and graph anomaly scoring. Orange and blue nodes denote the abnormal and normal nodes in a given graph. Firstly, a selected target node and two sampled subgraphs are encoded via a GNN encoder in the leftmost part. After this, two self-supervised objectives are introduced in the middle part, where $\mathcal{D}$ and $\mathcal{R}$ are shared discriminators and regressors. The GNN decoder are also shared across two graph views. Finally, in the rightmost part, we calculate the anomaly score for each node from two sources, where the anomalies can be easily detected (e.g., the orange node). }
\label{fig:framework}
\end{figure*}

\begin{definition}[Graph]
Given an attributed graph $\mathcal{G}=(\mathbf{X},\mathbf{A})$, we use $\mathbf{X} \in \mathbb{R}^{N \times D}$ and $\mathbf{A} \in \mathbb{R}^{N \times N}$ to denote the node features and graph adjacency matrix, where $N=|\mathcal{V}|$ and $\mathcal{V}=\{v_1, v_2, \dots, v_N\}$ is a set of nodes in the graph. Specifically, we use $\mathbf{x}_i \in \mathbb{R}^{D}$ to symbolize the feature of node $v_i$. Let $\mathcal{E}$ \hl{denotes} a set of graph edges, where $e_{ij}=\{v_i, v_j\} \in \mathcal{E}$ is an edge between node $v_i$ and $v_j$. The underlying graph structure is represented by a \hl{$N \times N$ square matrix $\mathbf{A}$}, where $\mathbf{A}_{ij}=1$ if $e_{ij}=\{v_i, v_j\}$, otherwise $\mathbf{A}_{ij}=0$. Particularly, the neighborhood of a node \hl{$v$} is defined as $\mathcal{N}(v)=\{u \in \mathcal{V}|e_{vu} \in \mathcal{E}\}$.
\end{definition}

\begin{definition}[Graph Neural Networks]
Given an attributed graph $\mathcal{G}=(\mathbf{X},\mathbf{A})$, a graph neural network $GNN(\cdot)$ aims to learn a local aggregation rule to map the original node features $\mathbf{X} = \{\mathbf{x}_1, \mathbf{x}_2, \dots, \mathbf{x}_N\} \in \mathbb{R}^{N \times D}$ to low-dimensional representations (i.e., embeddings) $\mathbf{H} = \{\mathbf{h}_1, \mathbf{h}_2, \dots, \mathbf{h}_N\} \in \mathbb{R}^{N \times D'}$.
\end{definition}

In this paper, we mainly focus on the problem of unsupervised anomaly detection on attributed graphs, which is defined below:

\begin{definition}[Unsupervised Graph Anomaly Detection]
Provided an attributed graph $\mathcal{G}=(\mathbf{X},\mathbf{A})$, we aim to learn a model $\mathcal{F}(\cdot): \mathbb{R}^{N \times D} \to \mathbb{R}^{N \times 1}$, which measures the degree of abnormality of a node in the graph by calculating its anomaly score without relying on any labelling information. Thus, the main task is to rank nodes according to their anomaly scores in a descending order, where anomalies can be easily detected based on this ranking list.
\end{definition}

\section{Methodology}
In this section, we present the overall framework of our proposed algorithm $\tool$ to detect node-level graph anomalies in an unsupervised manner. As shown in Figure \ref{fig:framework}, our method has three different components, including \textit{graph view sampling}, \textit{generative and contrastive self-supervised learning}, and \textit{graph anomaly scoring}. Firstly, we select a target node from the input graph, then we exploit the contextual information for this node. Specifically, we generate two associated graph views  by leveraging different \hl{augmentations}. After this, to fully utilize the rich node- and subgraph-level information to detect anomalies, we \hl{construct} two different self-supervised objectives, namely \textit{generative attribute reconstruction} and \textit{multi-view contrastive learning}. The former objective is inspired by the idea of graph auto-encoder (GAE) \cite{kipf2016variational}, which aims to reconstruct the feature vector of target node based on its neighboring attributive information. In such a way, if a selected target node is an anomaly, the attributive mismatch between it and its surrounding \hl{contexts} can be reflected as the regression loss between its reconstructed and original feature vector. Similar to but different from this node-level generative objective, we introduce another mixed-level contrastive objective \hl{to compare} a target node with its surrounding contexts directly on the embedding and structure space, which injects richer structural information during the discrimination. As a result, our model optimizes two self-supervised objectives that are closely related to the graph anomaly detection. During the inference, two scoring functions \hl{are} elaborately designed based on the aforementioned two objectives, which tend to assign \hl{attributive and structural} anomalies in a graph with higher anomaly scores.

In the rest of this section, we introduce the three core components of $\tool$ from Subsection \ref{subsec:sampling} to \ref{subsec:scoring}. In Subsection \ref{subsec:algorithm}, we present and analyse the training objective, algorithm, and its time complexity.

\subsection{Graph View Establishment for Anomaly Detection}\label{subsec:sampling}
Recent work in graph self-supervised learning suggests that the design of discrimination pairs is the key to allow graph encoders \hl{extracting} rich structural and attributive information \cite{velivckovic2018deep, hassani2020contrastive, hu2020strategies, you2020does}. Similar to the visual domain, graph self-supervised learning can be roughly divided into two categories: \hl{Generative-based and Contrastive-based.} As for the generative branch, existing work mainly lies on the attributive and structural auxiliary property prediction \cite{liu2021graph}, where the comparisons are typically placed on the same scale, such as "node v.s. node" (e.g., attribute regression) and "graph v.s. graph" (e.g., structure prediction). \hl{On the other hand,} contrastive learning could discriminate instances \hl{not only from} the same but also across different scales, such as "node v.s. graph" in \cite{hassani2020contrastive} and \cite{Jiao2020SubgraphCF}. However, not all of aforementioned discriminations are applicable to our task because graph anomaly detection and representation learning are two fundamentally different tasks. In graph anomaly detection, inspired by \cite{liuanomaly}, we conjecture that an anomaly is typically reflected as the mismatch between it and its surrounding contexts, which forms the foundation of our graph view construction.

In our method, to establish the connection between \hl{a target node and its surrounding contexts,} we propose two self-supervised learning objectives from different scales and spaces. Specifically, we first \hl{conduct the node-level discrimination, which reconstructs the feature vector of a target node by leveraging a GAE and then compares it with ground truth in the attribute space.}
% To answer reviewer 1's question: The reason to sample two subgraphs instead of three or more needs to be presented
% This part has also been modified to answer reviewer 2's question together with the last sentence on the previous paragraph: What are the connections between graph anomaly detection and so-called semi-global information
\hl{To inject richer structural information,} we further construct a mixed-level contrastiveness between a target node and its local subgraphs in the embedding and structure space, \hl{where sampling multiple views benefits our contrastive module exploring diverse semi-global (i.e., surrounding contextual) information during its discrimination \cite{tian2020contrastive}.}
Based on this, as shown in the leftmost part in Figure \ref{fig:framework}, we first sample a target node from the input graph, and then \hl{we sample two different views (local subgraphs)} around it by leveraging different graph augmentations.
% To answer reviewer 1's question: The reason to sample two subgraphs instead of three or more needs to be presented
\hl{Although it is possible to equip $\tool$ with more than two views, it may introduce redundant information depending on the choice of augmentations \cite{hassani2020contrastive, tosh2021contrastive} and thus degrading model performance.}

For our generative objective, the discrimination pair is the original and reconstructed target node. On the other hand, the target node and \hl{two sampled} graph views consist of two discrimination pairs \hl{in our contrastive objective.} Now we give a detailed explanation of the aforementioned processing steps:

\begin{enumerate}
	\item  \textbf{Target node sampling.} Since we mainly focus on detecting node-level anomalies in graphs, a target node needs to be sampled first. In this paper, we sample a target node from the given input graph via the uniform sampling without replacement.
	
	\item  \textbf{Graph view sampling.} %Augmentation plays a vital role in graph view sampling, which facilitates the proposed generative and contrastive objectives by introducing various discriminative contextual information. 
	\hl{Although several augmentations have been proposed on graphs,} such as node feature masking and edge modification \cite{Jin2021MultiScaleCS}, these augmentations are not applicable to graph anomaly detection because they introduce extra anomalies (e.g., modifying the underlying linkages and node features). Alternatively, in this paper, %we choose subgraph sampling as our augmentation approach to sample graph views centered at a target node with the fixed size $K$ by leveraging the random walk algorithm \cite{rwr_tong2006fast}, without violating the underlying graph semantic information.
	% To address reviewer 1's concern: The authors do not explicitly introduce how two subgraphs are sampled;
	% To answer reviewer 4's question: What is your intuition to using these subgraph sizes on 5.5.3?
	\hl{we leverage random walks with restart (RWR) \cite{rwr_tong2006fast} as augmentations, which avoid violating the underlying graph semantic information. Specifically, this approach samples graph views centered at a target node with the fixed size $K$, which controls the radius of surrounding contexts.}
	It worths noting that graph diffusion \cite{diffusion_klicpera2019diffusion} could also be a possible augmentation in our method to further injects the global information into our multi-view contrastiveness, which we leave in our future work.
	
	\item  \textbf{Graph view anonymization.} The target node in sampled graph views is anonymized (i.e., its features have been zeroed) to increase the difficulty of two predefined  self-supervised learning pretext tasks, which helps facilitate the model training \cite{liu2021graph, you2020graph}. In such a way, the raw attributive information of the target node will not contribute to its feature reconstruction, as well as the calculation of graph view embeddings. This mechanism prevents the information leakage and encourages the model to identify anomalies by relying on the contextual information.
\end{enumerate}

\subsection{Generative Learning with Attribute Reconstruction}\label{subsec:generative}
As suggested in \cite{dominant_ding2019deep}, the abnormality of an instance can be typically reflected as the degree of mismatch between its original and reconstructed information. Specifically, this type of mismatch can be quantified by the $\ell_2$-norm distance, where a higher distance denotes a higher reconstruction error, indicating that the given instance is more likely to be an anomaly. Among existing anomaly detection methods, deep autoencoder (AE) \cite{zhou2017anomaly} \hl{has} shown a strong performance. AE is a type of neural network that has been firstly introduced to learn latent representations in an unsupervised manner, which consists of two components: Deep encoder and decoder. Given an input feature vector $\mathbf{x}$, a typical AE can be defined as:
\begin{equation}
\mathbf{x'}=\operatorname{AE}(\mathbf{x})=f_{dec}\big(f_{enc}(\mathbf{x})\big),
\label{eq: autoencoder}
\end{equation}
where $\mathbf{x'}=f_{dec}(\mathbf{h})$ and $\mathbf{h}=f_{enc}(\mathbf{x})$ are deep \hl{decoder and encoder, respectively.} In above equation, $\mathbf{x'}$ is the reconstructed feature vector, and $\mathbf{h}$ denotes the latent representation of $\mathbf{x}$. The optimization objective of AE \hl{is to make} $\mathbf{x'}$ and $\mathbf{x}$ as close as possible, 
% which can be achieved by minimizing their $\ell_2$-norm distance:
% \begin{equation}
% dist = \lVert \mathbf{x' - x} \rVert^2_2,
% \label{eq: generative loss}
% \end{equation}
% which encourages the AE to learn latent invariant patterns among the inputs.
which can be achieved by minimizing their $\ell_2$-norm distance, \hl{i.e., $\lVert \mathbf{x' - x} \rVert^2_2$, to encourage AE to learn latent invariant patterns among inputs.}

%In graph anomaly detection, the abnormality of a node is typically reflected as the disparity between it and its neighbors (i.e., contextual information). 
\hl{However, in the context of graph anomaly detection,} a conventional AE only reconstructs the attributive information of a node and fails to take the underlying topological information into the consideration, which makes it unfavorable to \hl{this task.} To alleviate this limitation, as shown in the middle part of Figure \ref{fig:framework}, we construct \hl{a} GAE with two components: GNN-based encoder and decoder. \\

\noindent \textbf{GNN-based encoder.} Given two graph views $\mathcal{G}_t^{\phi_1}$ and $\mathcal{G}_t^{\phi_2}$ of a target node $v_t$, we resort to transform their high-dimensional node features to low-dimensional representations via a GNN encoder, which is formulated as:
\begin{equation}
\mathbf{H}_{\phi_i} = \operatorname{GNN}_{enc}(\mathbf{X}_{t}^{\phi_i}, \mathbf{A}_{t}^{\phi_i}),
\label{eq: encoder}
\end{equation}
where $\mathbf{H}_{\phi_i}$, $\mathbf{X}_{t}^{\phi_i}$, and $\mathbf{A}_{t}^{\phi_i}$ denote the node embedding matrix, node feature matrix, and adjacency matrix of $\mathcal{G}_t^{\phi_i}$, respectively. $\operatorname{GNN}_{enc}(\cdot)$ is the graph encoder, which consists of a $L$-layer GNN. Specifically, we define it as follows:
\begin{align}
\mathbf{m}_{\phi_i}^{(l)}[j,:] &= \operatorname{AGGREGATE}^{(l)}(\mathbf{H}_{\phi_i}^{(l-1)}[k,:]: v_k \in \mathcal{N}(v_j)) \nonumber, \\
\mathbf{H}_{\phi_i}^{(l)}[j,:] &= \operatorname{COMBINE}^{(l)}(\mathbf{H}_{\phi_i}^{(l-1)}[j,:], \mathbf{m}_{\phi_i}^{(l)}[j,:]) \label{eq: gnn},
\end{align}
where $\mathbf{H}_{\phi_i}^{(l)}[j,:]$ denotes the latent vector of $v_j$ in $\mathbf{H}_{\phi_i}^{(l)}$, which is the latent representation of $\mathcal{G}_t^{\phi_i}$ at the $l$-th layer. In particular, $\mathbf{H}_{\phi_i}=\mathbf{H}_{\phi_i}^{(L)}$ and $\mathbf{X}_{t}^{\phi_i}=\mathbf{H}_{\phi_i}^{(0)}$. In the above formulas, we use $\mathbf{m}_{\phi_i}^{(l)}[j,:]$ to denote the aggregated message of $v_j$ in $\mathcal{G}_t^{\phi_i}$ at the $l$-th layer. $\operatorname{AGGREGATE}^{(l)}(\cdot)$ and $\operatorname{COMBINE}^{(l)}(\cdot)$ are message aggregation and combination (a.k.a. transformation) functions in a typical GNN layer.

In this paper, for simplicity, we apply a one-layer GCN \cite{gcn_kipf2017semi} as our backbone graph encoder. In such a way, Equation \eqref{eq: encoder} can be specifically formulated as:
\begin{equation}
\mathbf{H}_{\phi_i} = \operatorname{GNN}_{enc}(\mathbf{X}_{t}^{\phi_i}, \mathbf{A}_{t}^{\phi_i}) = \sigma(\widehat{\mathbf{A}}_{t}^{\phi_i}\mathbf{X}_{t}^{\phi_i}\mathbf{W}_{enc}),
\label{eq: gcn encoder}
\end{equation}
where $\widehat{\mathbf{A}}_{t}^{\phi_i}=(\widetilde{\mathbf{D}}_{t}^{\phi_i})^{-\frac{1}{2}}\widetilde{\mathbf{A}}_{t}^{\phi_i}(\widetilde{\mathbf{D}}_{t}^{\phi_i})^{-\frac{1}{2}}$, $\widetilde{\mathbf{A}}_{t}^{\phi_i}=\mathbf{A}_{t}^{\phi_i}+\mathbf{I}$, and $\widetilde{\mathbf{D}}_{t}^{\phi_i}[i,i]=\sum_{j}\widetilde{\mathbf{A}}_{t}^{\phi_i}[i,j]$. Specifically, $\sigma(\cdot)$ denotes a non-linear activation function (e.g., ReLU), and $\mathbf{W}_{enc}$ is a trainable parameter matrix in our graph encoder. \\

\noindent \textbf{GNN-based decoder.} Similarly, we build our graph decoder with a single GCN layer, which is slightly different from Equation \eqref{eq: gcn encoder}. During the attribute reconstruction, we take node embedding matrix as the input and then reconstruct the node features accordingly:
\begin{equation}
\widehat{\mathbf{X}}_{t}^{\phi_i} = \operatorname{GNN}_{dec}(\mathbf{H}_{\phi_i}, \mathbf{A}_{t}^{\phi_i})= \sigma(\widehat{\mathbf{A}}_{t}^{\phi_i}\mathbf{H}_{\phi_i}\mathbf{W}_{dec}),
\label{eq: gcn decoder}
\end{equation}
where $\mathbf{W}_{dec}$ is a trainable parameter matrix in our graph decoder. \\

\noindent \textbf{Generative graph anomaly detection.} By aggregating the neighboring information, the attributive reconstruction of $\mathcal{G}_t^{\phi_1}$ and $\mathcal{G}_t^{\phi_2}$ is mainly based on the attributive information of local surrounding subgraphs centred at $v_t$. In this paper, we anonymize the target node $v_t$ in its two graph views to enforce its attributive reconstruction purely based on the contextual information. As we mentioned, this \hl{schema} better reflects node-level anomalies in the attribute space. Thus, we propose to minimize the Mean Squared Error (MSE) between target node's original and reconstructed features in two graph views, as shown in Figure \ref{fig:framework}.
% \begin{equation}
% \mathcal{L}^{1}_{gen}=\frac{1}{N}\sum_{i=1}^N(\widehat{\mathbf{X}}_{i}^{\phi_1}[-1,:], \mathbf{x}_{i})^2,
% \label{eq: genloss1}
% \end{equation}
% \begin{equation}
% \mathcal{L}^{2}_{gen}=\frac{1}{N}\sum_{i=1}^N(\widehat{\mathbf{X}}_{i}^{\phi_2}[-1,:], \mathbf{x}_{i})^2,
% \label{eq: genloss2}
% \end{equation}
\begin{equation}
\hl{\mathcal{L}^{j}_{gen}=\frac{1}{N}\sum_{i=1}^N(\widehat{\mathbf{X}}_{i}^{\phi_j}[-1,:], \mathbf{x}_{i})^2, \  j \in \{1,2\},}
\label{eq: genloss 1 and 2}
\end{equation}
where $\mathbf{x}_{i}$ is the feature vector of a target node $v_i$, and $\widehat{\mathbf{X}}_{i}^{\phi_j}[-1,:]$ denotes the reconstructed feature vector of anonymized target node $v_i$ in its $j$-th graph view. Specifically, $\widehat{\mathbf{X}}_{i}^{\phi_j}$ can be obtained via Equation \eqref{eq: gcn encoder} and \eqref{eq: gcn decoder}. %Finally, we have our final generative objective by combining Equation \eqref{eq: genloss1} and \eqref{eq: genloss2}:
\hl{Finally, we have our generative objective by combining $\mathcal{L}^{j}_{gen}$ in Equation \eqref{eq: genloss 1 and 2}:}
\begin{equation}
\mathcal{L}_{gen}=\frac{1}{2}(\mathcal{L}^{1}_{gen} + \mathcal{L}^{2}_{gen}).
\label{eq: genloss}
\end{equation}

\subsection{Multi-View Contrastive Learning}\label{subsec:contrastive}
As we mentioned before, \hl{node anomalies} are typically reflected as the mismatch between \hl{nodes and their surrounding contexts.} Our generative module identifies anomalies \hl{in} the attribute space with the help of our GNN encoder and decoder, \hl{but} the structural information has not been directly utilized.
% To overcome this limitation and inject richer structural information when discriminating a target node with its surrounding contexts, we propose a multi-view contrastive module, which contrasts a target node with two associated graph views directly.
To overcome this limitation and inject richer \hl{structural information,} we propose a multi-view contrastive module, which contrasts a target node with two associated graph views directly. 
% Different from the generative objective where the discrimination is placed on the node-level and attribute space, our multi-view contrastive learning mixes different graph topological levels, which discriminates the representation of a target node with its local subgraph representations on the embedding and structural space, emphasizing more on graph topological information. 
Different from the generative objective where the discrimination is placed on the node-level and \hl{in the} attribute space, our multi-view contrastive learning mixes different graph topological \hl{scales,} which discriminates the representation of a target node with its local subgraph representations in the embedding and structural space, emphasizing more on \hl{semi-global information.}
As illustrated in Figure \ref{fig:framework}, our contrastive module mainly consists of three different components: Graph encoder, readout module, and contrastive module. \\

\noindent \textbf{GNN-based encoder.} Our contrastive module takes the node feature vector and matrices of a selected target node and two associated graph views as the input, where the underlying graph encoder shares the same parameters with the generative module. The encoding of two graph views has been formulated in Equation \eqref{eq: gcn encoder}, while the transformation of target node feature vector follows a different formula:
\begin{equation}
\mathbf{h}_t = \sigma(\mathbf{x}_{t}\mathbf{W}_{enc}),
\label{eq: gcn encoder on target node}
\end{equation}
where $\mathbf{h}_t$ and $\mathbf{x}_t$ are embedding and feature vectors of the target node $v_t$. It is worth noting that a single target node has no underlying graph structure, so the graph adjacency matrix in Equation \eqref{eq: gcn encoder} is discarded in \hl{Equation \eqref{eq: gcn encoder on target node}.} Thus, Equation \eqref{eq: gcn encoder on target node} is equivalent to a non-linear mapping of $\mathbf{x}_t$, where $\mathbf{W}_{enc}$ is shared with Equation \eqref{eq: gcn encoder}. \\

\noindent \textbf{Readout module.} Since we aim to contrast $\mathbf{h}_t$ with the surrounding subgraphs directly, we are motivated to design a readout module \hl{to generate} two semi-global (subgraph-level) representations based on $\mathbf{H}_{\phi_1}$ and $\mathbf{H}_{\phi_2}$, as shown in the middle part of Figure \ref{fig:framework}.

In general, several readout functions are commonly used to generate graph-level representations based on node-level embeddings, such as average and differentiable pooling \cite{hassani2020contrastive, ying2018hierarchical}. For simplicity, we adopt the average pooling in this paper, which can be formulated as:
\begin{equation}
\mathbf{g}_{\phi_i} = \frac{1}{K}\sum_{j=1}^{K}\mathbf{H}_{\phi_i}[j,:],
\label{eq: readout}
\end{equation}
where $\mathbf{g}_{\phi_i}$ denotes the graph-level embedding vector of view $\mathcal{G}_t^{\phi_i}$. $K$ is the number of nodes in a graph view. \\

\noindent \textbf{Contrastive module.} Given %the embedding vector of $v_t$,
\hl{$\mathbf{h}_t$}, $\mathbf{g}_{\phi_1}$, and $\mathbf{g}_{\phi_2}$, we aim to discriminate them pairwise. Specifically, we first formulate the positive and negative pairs of $v_t$ as follows:
\begin{equation}
P_t^{\phi_i} = (\mathbf{h}_t, \mathbf{g}_{\phi_i}),
\label{eq: positive}
\end{equation}
\begin{equation}
\widetilde{P}_t^{\phi_i} = (\mathbf{h}_t, \widetilde{\mathbf{g}}_{\phi_i}),
\label{eq: negative}
\end{equation}
where $P_t^{\phi_i}$ and $\widetilde{P}_t^{\phi_i}$ are positive and negative pairs of the selected target node $v_t$. In Equation \eqref{eq: negative}, $\widetilde{\mathbf{g}}_{\phi_i}$ denotes the negative samples, which are the representations of randomly cropped subgraphs from $\mathcal{G}$ and different from $\mathcal{G}_{t}^{\phi_1}$ and $\mathcal{G}_{t}^{\phi_2}$.

To contrast two elements in positive and negative pairs, we design a discriminator based on the bilinear transformation \cite{liuanomaly}, where the discrimination scores of $P_t^{\phi_i}$ and $\widetilde{P}_t^{\phi_i}$ are defined as follows:
\begin{equation}
s_t^{\phi_i}=\sigma(\mathbf{h}_{t}\mathbf{W}_{s}\mathbf{g}_{\phi_i}^\mathsf{T}),
\label{eq: positive discriminate}
\end{equation}
\begin{equation}
\widetilde{s}_t^{\phi_i}=\sigma(\mathbf{h}_{t}\mathbf{W}_{s}\widetilde{\mathbf{g}}_{\phi_i}^\mathsf{T}),
\label{eq: negative discriminate}
\end{equation}
where $s_t^{\phi_i}$ and $\widetilde{s}_t^{\phi_i}$ are contrastive discrimination scores of the pairs $P_t^{\phi_i}$ and $\widetilde{P}_t^{\phi_i}$. In the above equations, $\mathbf{W}_s \in \mathbb{R}^{D' \times D'}$ is a learnable scoring matrix, which measures the similarity between two input vectors. Particularly, we resort to use the sigmoid function as our non-linear transformation $\sigma(\cdot)$ in above two equations to ensure our discrimination scores fall within the range $[0,1]$. \\

\noindent \textbf{Multi-view contrastive graph anomaly detection.} Conceptually, graph anomalies should be different from their \hl{surrounding} contexts on both attributive and topological perspectives. Thus, we conjecture that $s_t^{\phi_i}$ should be significantly larger than $\widetilde{s}_t^{\phi_i}$, which indicates that for most of normal nodes in $\mathcal{G}$, target node representations (e.g., $\mathbf{h}_t$) should share more similarities with their surrounding contexts (e.g., $\mathbf{g}_{\phi_i}$) than other subgraphs (e.g., $\widetilde{\mathbf{g}}_{\phi_i}$).

Motivated by this, we form our multi-view contrastive objectives based on the Jensen-Shannon divergence \cite{velivckovic2018deep}, which maximizes the agreement between a target node and its surrounding contexts:
% \begin{equation}
% \mathcal{L}_{con}^{1}=-\frac{1}{2N}\sum_{i=1}^{N}(log(s_i^{\phi_1})+log(1-\widetilde{s}_i^{\phi_1})),
% \label{eq: conloss1}
% \end{equation}
% \begin{equation}
% \mathcal{L}_{con}^{2}=-\frac{1}{2N}\sum_{i=1}^{N}(log(s_i^{\phi_2})+log(1-\widetilde{s}_i^{\phi_2})),
% \label{eq: conloss2}
% \end{equation}
\begin{equation}
\hl{\mathcal{L}_{con}^{j}=-\frac{1}{2N}\sum_{i=1}^{N}(\log(s_i^{\phi_j})+\log(1-\widetilde{s}_i^{\phi_j})), \ j \in \{1,2\},}
\label{eq: conloss 1 and 2}
\end{equation}
% where $\mathcal{L}_{con}^{1}$ and $\mathcal{L}_{con}^{2}$ denote the contrastive loss of graph view 1 and 2 across all nodes in $\mathcal{G}$. By combining Equation \eqref{eq: conloss1} and \eqref{eq: conloss2}, we have our final contrastive objective:
where \hl{$\mathcal{L}_{con}^{j}$} denotes the contrastive loss of graph view $j$ across all nodes in $\mathcal{G}$. By combining \hl{$\mathcal{L}_{con}^{j}$ in Equation \eqref{eq: conloss 1 and 2},} we have our final contrastive objective:
\begin{equation}
\mathcal{L}_{con}=\frac{1}{2}(\mathcal{L}^{1}_{con} + \mathcal{L}^{2}_{con}).
\label{eq: conloss}
\end{equation}

\subsection{Graph Anomaly Scoring}\label{subsec:scoring}
Until now, we have introduced two different self-supervised discrimination schemes for graph anomaly detection by leveraging our generative and contrastive modules to compare nodes with their contextual information. As the majority of nodes in $\mathcal{G}$ are not anomalies, a well-trained graph encoder and decoder are expected to map the feature vector of a normal node to an appropriate latent space and vice versa. For anomalies in $\mathcal{G}$, their embeddings and reconstructed features are likely to be distorted because of their attributive or structural abnormalities.

For the generative learning, the attributive reconstruction of an anonymized target node is purely based on its local (i.e., neighbouring) contextual information, where the degree of mismatch between the reconstructed and original feature vector is an ideal metric to measure the abnormality of a node. In this paper, we adopt the $\ell_2$-norm distance %described in Equation \eqref{eq: generative loss} 
as the generative anomaly scoring function. For a node $v_i \in \mathcal{V}$, we define this function as follows:
\begin{equation}
f_{gen}(v_i)=\frac{1}{2}\sum_{j=1}^{2}\left(\hl{\delta^1}(\lVert \widehat{\mathbf{X}}_{i}^{\phi_j}[-1,:] - \mathbf{x}_i \rVert^2_2)\right),
\label{eq: generative socre}
\end{equation}
where $f_{gen}(v_i)$ denotes the generative anomaly score of $v_i$. In above formula, $\widehat{\mathbf{X}}_{i}^{\phi_j}[-1,:]$ is the reconstructed feature vector of $v_i$, which is calculated in Equation \eqref{eq: gcn decoder}. Specifically, \hl{$\delta^1$} denotes a scaler, which scales a $\ell_2$-norm distance to the range of $[0,1]$. In such a way, if $v_i$ is an anomaly, $f_{gen}(v_i)$ is expected to be close to 1, otherwise this score should be close to 0.

However, our generative learning focuses only the node-level discrimination on the attribute space, where the graph topological information has not been directly utilized. \hl{To overcome this limitation%and inject richer semi-global information during the discrimination
,} our contrastive module discriminates a target node with the surrounding subgraphs directly on the embedding space, where the discrimination scores in Equation \eqref{eq: positive discriminate} and \eqref{eq: negative discriminate} can be naturally combined to form an abnormality metric. Inspired by \cite{liuanomaly}, for a node $v_i \in \mathcal{V}$, we define our contrastive anomaly scoring function as follows:
\begin{equation}
f_{con}(v_i)=\frac{1}{2}\sum_{j=1}^{2}\hl{\delta^2}\left(\widetilde{s}_i^{\phi_j}-s_{i}^{\phi_j}\right),
\label{eq: contrastive socre}
\end{equation}
where $f_{con}(v_i)$ denotes the contrastive anomaly score of $v_i$. Specifically, $s_{i}^{\phi_j}$ and $\widetilde{s}_i^{\phi_j}$ are positive and negative contrastive discrimination scores defined in Equation \eqref{eq: positive discriminate} and \eqref{eq: negative discriminate}. If $v_i$ is a normal node, $s_{i}^{\phi_j}$ is expected to be close to 1, and $\widetilde{s}_i^{\phi_j}$ is likely to be close to 0. Otherwise, if $v_i$ is an anomaly, $s_{i}^{\phi_j}$ and $\widetilde{s}_i^{\phi_j}$ are expected to be close to 0.5 because of the mismatch between $v_i$ and its surrounding subgraphs. Thus,  \hl{$\widetilde{s}_i^{\phi_j}-s_{i}^{\phi_j}$ falls} in the range of $[-1, 0]$. By introduce a scaler \hl{$\delta^2$}, our final contrastive anomaly scores will be scaled to the range of $[0,1]$.

By combining aforementioned two anomaly scoring functions, we have our final graph anomaly scoring function to estimate the abnormality of a node $v_i \in \mathcal{V}$:
\begin{equation}
f(v_i)=\alpha f_{con}(v_i)+\beta f_{gen}(v_i),
\label{eq: final score}
\end{equation}
where $\alpha$ and $\beta$ are two tunable balancing factors to weight the importance of contrastive and generative scoring functions. In practice, as suggested in \cite{liuanomaly}, we may have to repeat this calculation $R$ times to obtain a statistical stable anomaly score. This is because two sampled graph views $\mathcal{G}_{t}^{\phi_1}$ and $\mathcal{G}_{t}^{\phi_2}$ are partial observations on $v_t$'s contextual information, which may be insufficient to estimate the abnormality of $v_t$.

\subsection{Model Optimization and Algorithm}\label{subsec:algorithm}
By combining our generative and contrastive objectives in Equation \eqref{eq: genloss} and \eqref{eq: conloss}, we have our final optimization goal:
\begin{equation}
\mathcal{L}=\alpha\mathcal{L}_{con}+\beta\mathcal{L}_{gen},
\label{eq: obj_func}
\end{equation}
where $\mathcal{L}$ is the training loss to be minimized. $\alpha$ and $\beta$ are two balancing factors, which are the same as in Equation \eqref{eq: final score} to balance the importance of two self-supervised modules.

\begin{algorithm}[t]
	\caption{The Proposed $\tool$ Algorithm}
	\label{algo: overall algorithm}
    \textbf{Input}: Attributed graph \hl{$\mathcal{G}$}; Maximum training epochs $E$; Batch size $B$; Number of evaluation rounds $R$. \\
    % \textbf{Output}: Pre-trained GNN encoder $\operatorname{GNN}_{enc}$ and GNN decoder $\operatorname{GNN}_{dec}$; Graph anomaly scoring function $f(\cdot)$.
    \textbf{Output}: \hl{Graph anomaly scoring function $f(\cdot)$.} \\ \vspace{-4mm}
    \begin{algorithmic}[1]
		\STATE Randomly initialize the trainable parameters \hl{$\mathbf{W}_{enc}$, $\mathbf{W}_{dec}$, and $\mathbf{W}_{s}$;}
		\STATE $//$ {\it Training stage}
		\FOR{$e \in 1,2,\cdots,E$}
			\STATE $\mathcal{B} \leftarrow$ Randomly split $\mathcal{V}$ into batches with size $B$;
			\FOR{batch $b=(v_{1},\cdots,v_{B}) \in \mathcal{B}$}
			\STATE Sample two graph views for each of node in $b$, i.e., $(\mathcal{G}_1^{\phi_1},\cdots,\mathcal{G}_B^{\phi_1})$ and $(\mathcal{G}_1^{\phi_2},\cdots,\mathcal{G}_B^{\phi_2})$;
			\STATE Calculate node and associated graph view embeddings via Eq. \eqref{eq: gcn encoder on target node} and \eqref{eq: gcn encoder};
			\STATE Calculate the reconstructed feature vectors for nodes in $b$ via Eq. \eqref{eq: gcn decoder};
% 			\STATE Calculate the generative loss $\mathcal{L}_{gen}$ via Eq. \eqref{eq: genloss};
% 			\STATE Calculate the contrastive loss $\mathcal{L}_{con}$ via Eq. \eqref{eq: conloss};
% 			\STATE Calculate the overall training loss $\mathcal{L}$ via Eq. \eqref{eq: obj_func};
            \STATE \hl{Calculate the loss $\mathcal{L}$ via Eq. \eqref{eq: genloss}, \eqref{eq: conloss}, and \eqref{eq: obj_func};}
			\STATE Back propagation and update trainable parameters \hl{$\mathbf{W}_{enc}$, $\mathbf{W}_{dec}$, and $\mathbf{W}_{s}$;}
			\ENDFOR
		\ENDFOR
		\STATE $//$ {\it Inference stage}
		\FOR{$v_i \in \mathcal{V}$}
		\FOR{evaluation round $r \in 1,2,\cdots,R$}
		\STATE \hl{Calculate $f(v_i)$ via Eq. \eqref{eq: generative socre}, \eqref{eq: contrastive socre}, and \eqref{eq: final score};}
% 		\STATE Calculate $f_{con}(v_i)$ via Eq. \eqref{eq: contrastive socre};
% 		\STATE Calculate $f(v_i)$ via Eq. \eqref{eq: final score};
		\ENDFOR
		\STATE Average $f(v_i)$ over $R$ evaluation rounds and output the final anomaly score for $v_i$;
		\ENDFOR
	\end{algorithmic}
\end{algorithm}

The overall procedures of the proposed $\tool$ are summarized in Algorithm \ref{algo: overall algorithm}. Firstly, we sample a batch of nodes from the input graph. For each of node, we generate two graph views. After this, nodes in a batch and associated graph views are encoded via a trainable GNN encoder. To calculate the generative loss, the node embeddings of graph views are decoded via a GNN decoder, where we aim to reconstruct the feature vector of anonymized target node in each graph view. Then, we compare the reconstructed feature vector of a node with its original feature vector. Simultaneously, the node embeddings of graph views are aggregated to generate view representations, which are then compared with the target node embeddings to calculate the contrastive loss. Finally, by combining two different objectives, we calculate the final training loss, where different trainable parameters can be updated via backpropagation. During the inference, the anomaly score of each node in $\mathcal{G}$ will be repetitively calculated $R$ times with different graph views sampled each time, which ensures the final anomaly scores are statistically stable.

\subsubsection{Complexity Analysis}
\hl{In this subsection, we analyse the time complexity of the proposed $\tool$ algorithm. For graph view sampling on a selected target node $v_t$, the time complexity of our RWR-based approach is $\mathcal{O}(\eta K)$, where $\eta$ denotes the average node degree in the graph. Given a graph view $\mathcal{G}_t^{\phi_i}$ with $K$ nodes, the time complexity of the proposed two self-supervised learning modules is $\mathcal{O}(K^2)$. Thus, considering a graph with $N$ nodes, the training complexity of our algorithm is $\mathcal{O}(N K(\eta+K))$. During the inference, our graph anomaly scoring has a constant time complexity. Considering we have $R$ evaluation rounds, the time complexity of our algorithm is $\mathcal{O}(R N K(\eta+K))$ when considering both model training and inference.}

\section{Experimental Study} \label{sec:experiment}
In this section, we conduct comprehensive experiments on six real-world benchmark datasets to demonstrate the effectiveness of our proposed $\tool$ model. We compare our method with the state-of-the-art anomaly detection and self-supervised learning methods, and follow their configurations to carry out our experiments for a fair comparison. We conduct ablation study and parameter sensitivity experiments to further investigate the property of $\tool$.

\begin{table}[ht]
	\centering
	\caption{The statistics of the datasets. The upper two datasets are social networks, and the remainders are citation networks.}
	\begin{tabular}{@{}c|c|c|c|c@{}}
		\toprule
		\textbf{Dataset}       &  \textbf{Nodes} & \textbf{Edges} & \textbf{Features} & \textbf{Anomalies} \\ 
		\midrule
		\textbf{BlogCatalog} \cite{tang2009relational} & 5,196           & 171,743        & 8,189             & 300                \\
		\textbf{Flickr} \cite{tang2009relational}      & 7,575           & 239,738        & 12,407            & 450                \\
		\textbf{ACM}  \cite{tang2008arnetminer}        & 16,484          & 71,980         & 8,337             & 600                \\ 
		\textbf{Cora}  \cite{sen2008collective}        & 2,708           & 5,429          & 1,433             & 150                \\
		\textbf{CiteSeer}  \cite{sen2008collective}    & 3,327           & 4,732          & 3,703             & 150                \\
		\textbf{Pubmed}   \cite{sen2008collective}     & 19,717          & 44,338         & 500               & 600                \\
		\bottomrule
	\end{tabular}
	\label{table:dataset}
\end{table}

\subsection{Dataset Description}
To evaluate the performance of $\tool$ and the competitors, six real-world graph datasets, including two social network datasets and four citation network datasets, are utilized as benchmarks in our experiments. We provide the description of these datasets as follows:
% Dataset statistics are summarized in Table \ref{table:dataset}.

\begin{itemize}
	
	\item \textbf{Social Networks}. Blogcatalog and Flickr\footnote{http://socialcomputing.asu.edu/pages/datasets} \cite{tang2009relational} are two social network datasets whose data \hl{are} collected from the blog sharing website BlogCatalog and the image sharing website Flickr, respectively. In these social network datasets, each node represents a user of websites, and each link indicates the following relationships between two users. The personalized contents (e.g., posting blogs or sharing photos with tag descriptions) of users are extracted as node features. 
	
	\item \textbf{Citation Networks}. Cora, CiteSeer, Pubmed\footnote{http://linqs.cs.umd.edu/projects/projects/lbc} \cite{sen2008collective} and ACM \cite{tang2008arnetminer} are four public citation graph datasets. The data is collected from the corresponding publication databases. In these graphs, each node is a published paper, while each edge denotes a citation relationship between two papers. The text contents of each paper are treated as its node features.
	
\end{itemize}

\begin{table*}[htbp]
	\small
	\centering
	\caption{Anomaly detection performance (i.e., AUC values) on six benchmark datasets. The best performance on each dataset is in bold.}
	{
    	\begin{tabular}{p{75 pt}<{\centering}|p{45 pt}<{\centering}p{45 pt}<{\centering}p{45 pt}<{\centering}p{45 pt}<{\centering}p{45 pt}<{\centering}p{45 pt}<{\centering}}%{lcccccc}
    		\toprule
    		Method      & BlogCatalog     & Flickr          & ACM             & Cora            & CiteSeer        & PubMed          \\
    		\midrule
			AMEN \cite{amen_perozzi2016scalable}        & 0.6392          & 0.6573          & 0.5626          & 0.6266          & 0.6154          & 0.7713          \\
			Radar \cite{radar_li2017radar}      & 0.7401          & 0.7399          & 0.7247          & 0.6587          & 0.6709          & 0.6233          \\
			ANOMALOUS \cite{anomalous_peng2018anomalous}  & 0.7237          & 0.7434          & 0.7038          & 0.5770          & 0.6307          & 0.7316          \\
			\midrule
			DOMINANT \cite{dominant_ding2019deep}   & 0.7468          & 0.7442          & 0.7601          & 0.8155          & 0.8251          & 0.8081          \\
			DGI  \cite{dgi_velickovic2019deep}       & 0.5827          & 0.6237          & 0.6240          & 0.7511          & 0.8293          & 0.6962          \\
			\hl{CoLA  \cite{liuanomaly}}      & \hl{0.7854}          & \hl{0.7513}          & \hl{0.8237}          & \hl{0.8779}          & \hl{0.8968}          & \hl{0.9512}          \\
    		\midrule
    		\tool       & \textbf{0.8184} & \textbf{0.7966} & \textbf{0.8538} & \textbf{0.9130} & \textbf{0.9136} & \textbf{0.9672} \\
    		\bottomrule
    	\end{tabular}
    }
\label{table:overall}
\end{table*}

\begin{figure*}[htbp]
\centering
\subfigure[BlogCatalog]{
\includegraphics[width=5cm]{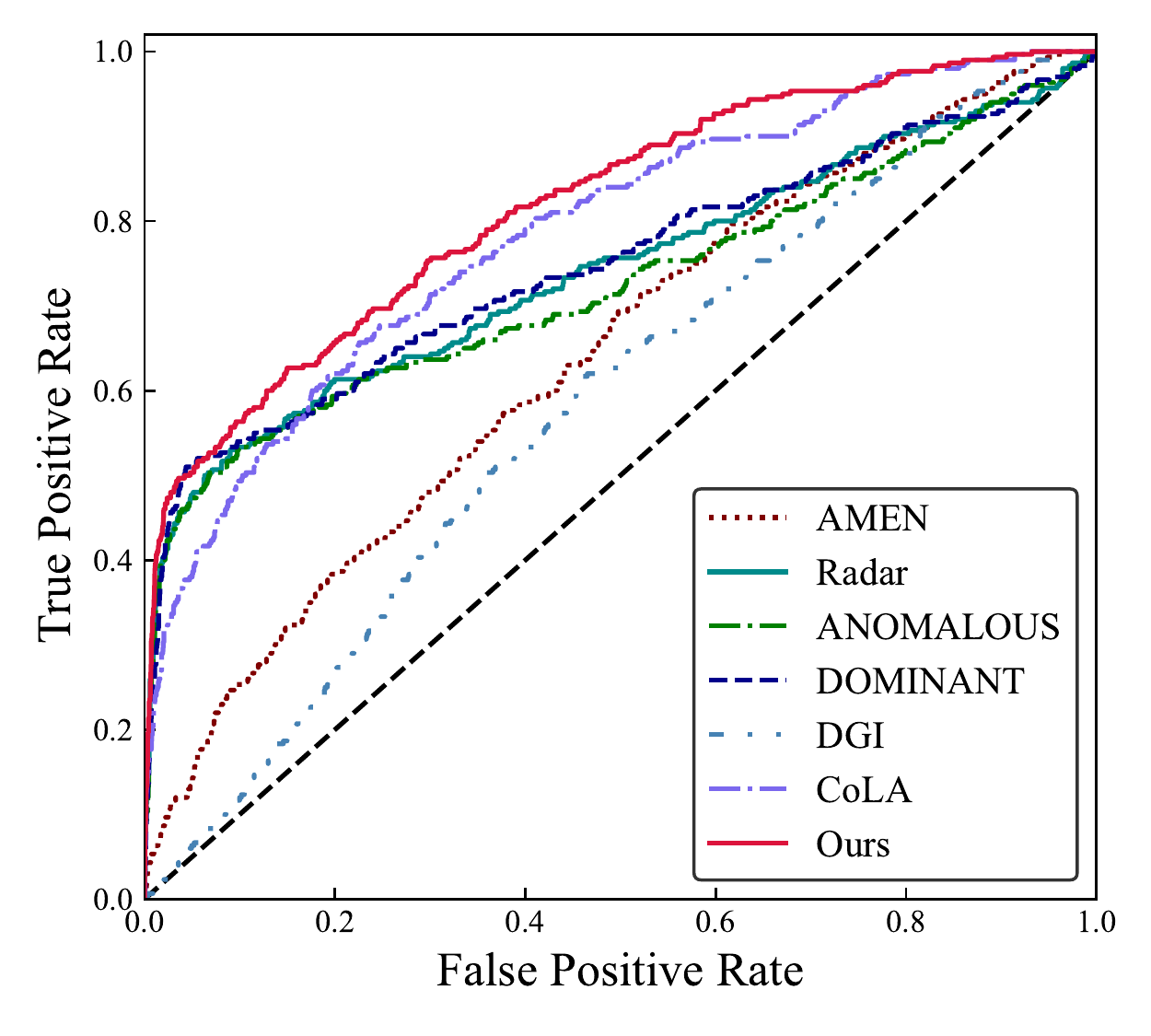}
}
\quad
\subfigure[Flickr]{
\includegraphics[width=5cm]{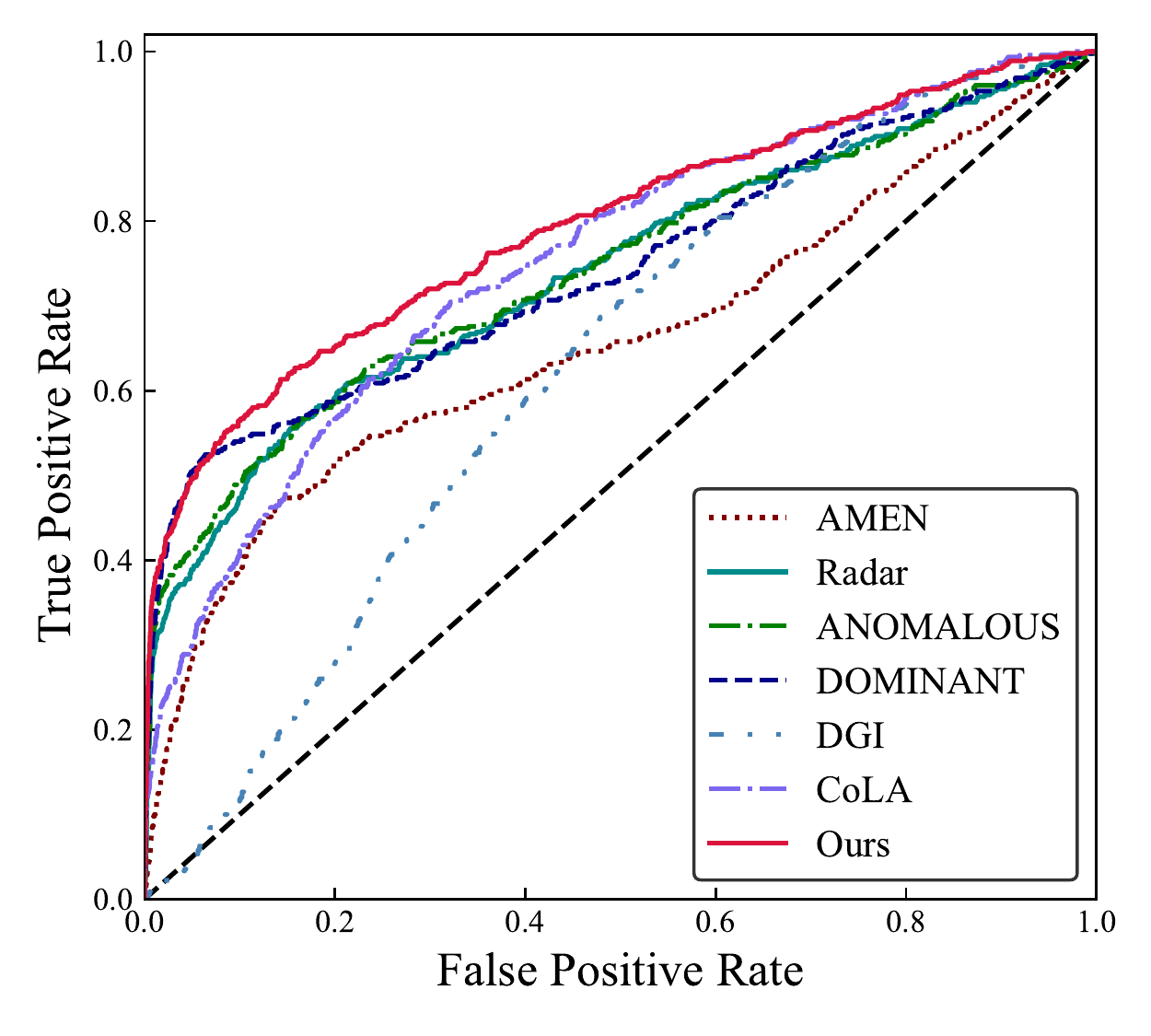}
}
\quad
\subfigure[ACM]{
\includegraphics[width=5cm]{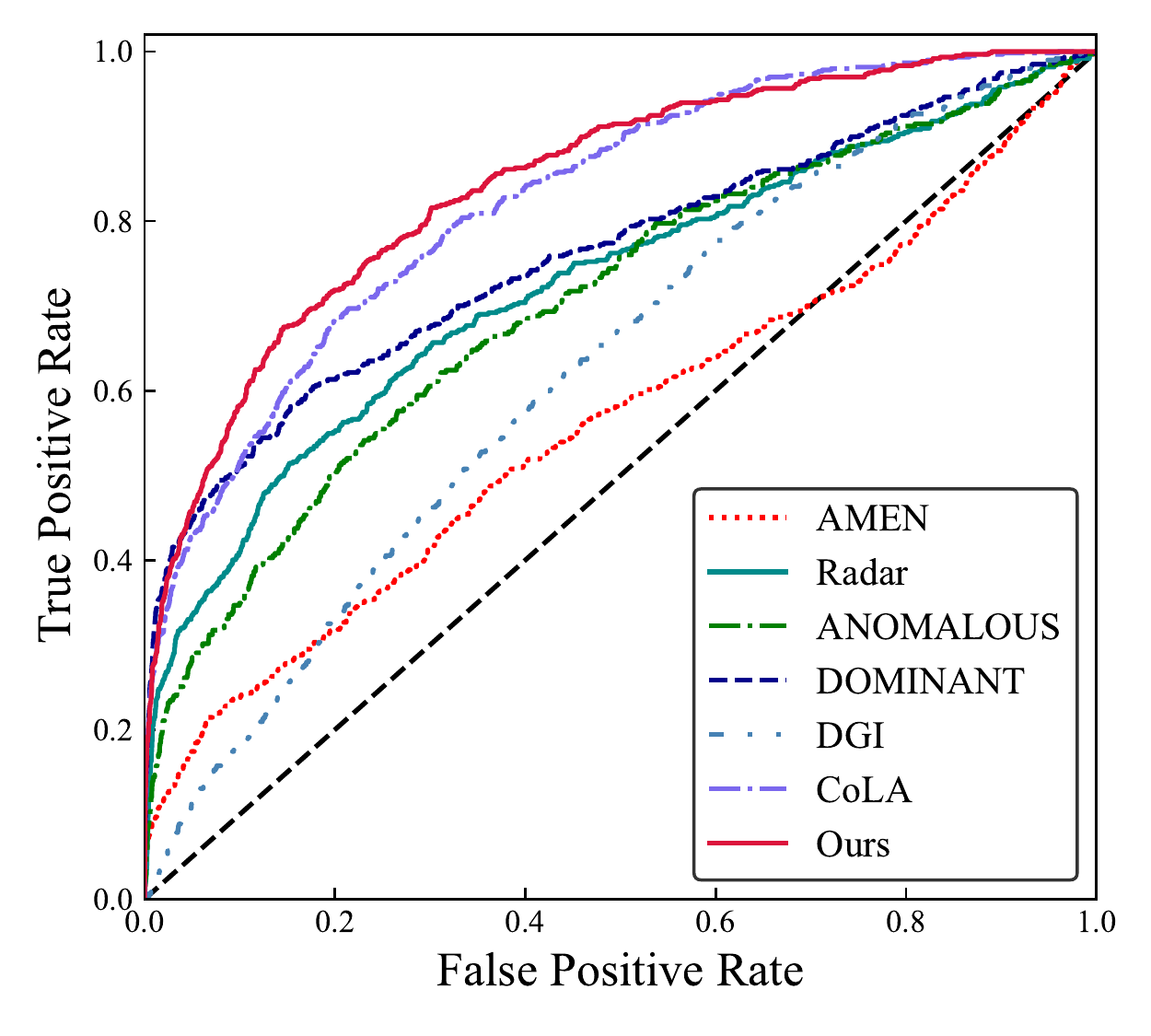}
}

\subfigure[Cora]{
\includegraphics[width=5cm]{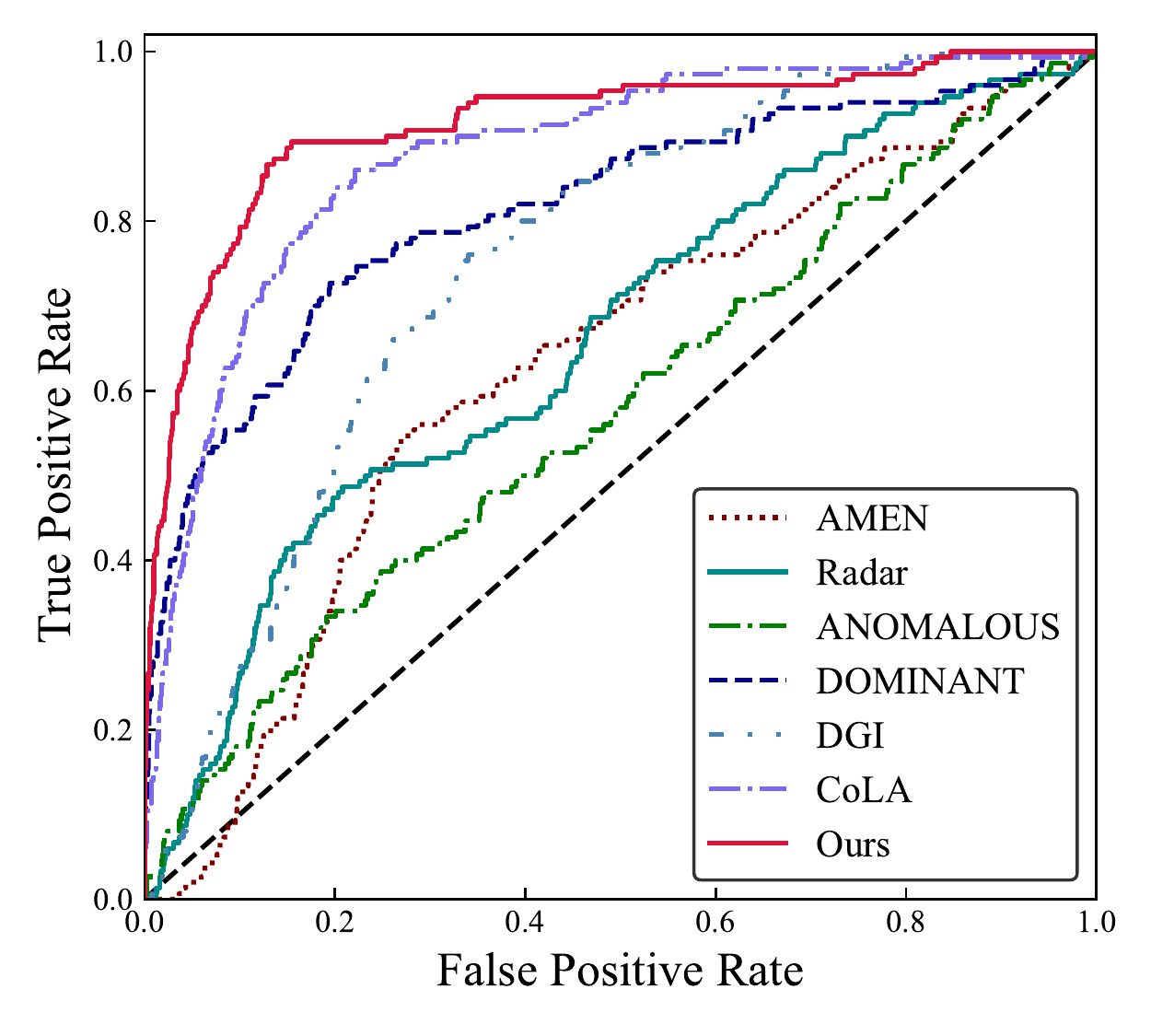}
}
\quad
\subfigure[CiteSeer]{
\includegraphics[width=5cm]{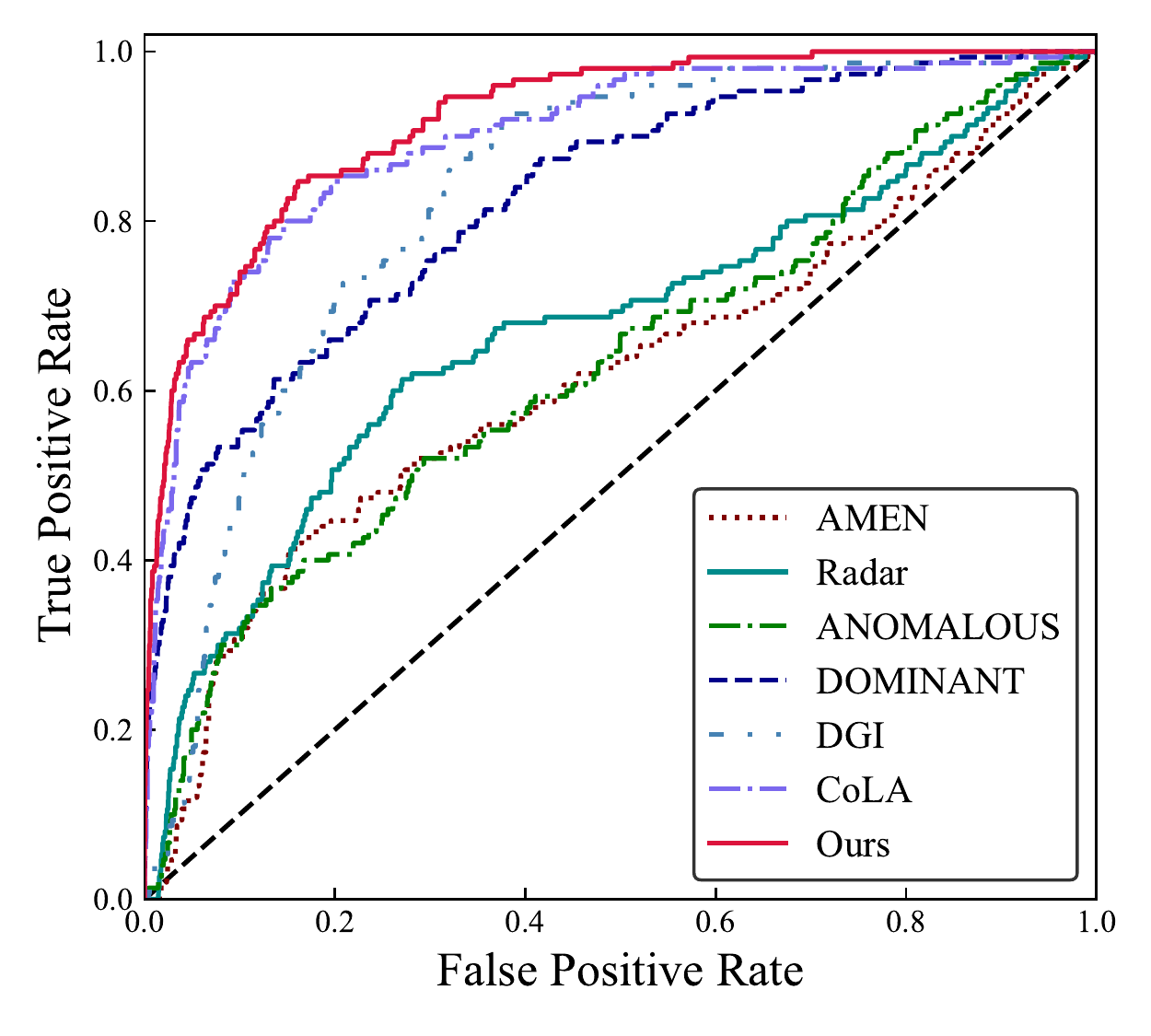}
}
\quad
\subfigure[Pubmed]{
\includegraphics[width=5cm]{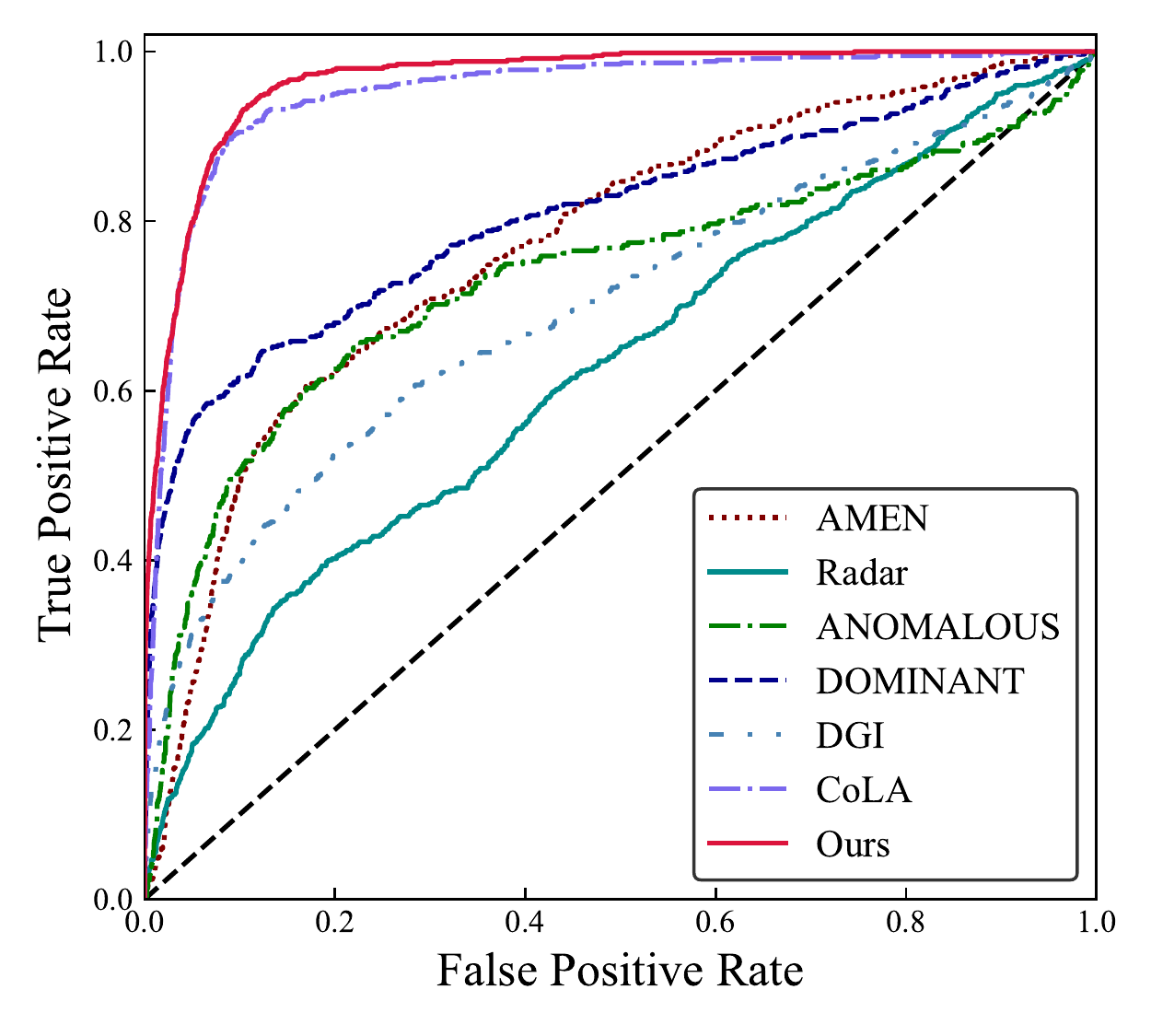}
}
\caption{ROC curves on \hl{six} benchmark datasets. The larger the area under the curve, the better the performance of graph anomaly detection. "Ours" in the figure legends denotes the proposed method $\tool$.}
\label{fig:roc}
\end{figure*}

Since no ground-truth anomalies are included in these datasets, we need to inject synthetic anomalous data into the original data manually. \hl{To compare fairly, we follow the anomaly generation strategy used in \cite{dominant_ding2019deep} and \cite{liuanomaly}.
% Equal numbers of structural and contextual anomalies are injected into the six datasets, and the total numbers of anomalies are also summarized in Table \ref{table:dataset} together with the dataset statistics.
Specifically, when injecting attributive anomalies, we select $N_a$ nodes and then replace their features with randomly selected distant nodes' features. After this, we further inject equal number of structural anomalies into six datasets by selecting $N_{a}'$ nodes and making them fully connected. We repeat this process $q$ times such that $N_a = q \times N_{a}'$. The total numbers of anomalies are also summarized in Table \ref{table:dataset} together with the dataset statistics.}

\subsection{Experimental Setup} \label{subsec:exp_setup}

We illustrate the experimental setup in this subsection, including baseline methods, evaluation metrics, and parameter settings.

\textbf{Baselines.} We compare $\tool$ against five state-of-the-art anomaly detection and self-supervised learning methods. We briefly introduce these methods as follows:

\begin{itemize} 
	\item \textbf{AMEN} \cite{amen_perozzi2016scalable} detects anomalies via ego-network analysis. It evaluates the correlation of attributes among different nodes within a ego-network to discriminate anomalous information. %{\footnote{https://github.com/phanein/amen}
	
	\item \textbf{Radar} \cite{radar_li2017radar} leverages residual analysis to identify the anomalies in graphs. It considers the residuals of attribute information and the coherence information with graph to detect anomalies. %\footnote{http://people.virginia.edu/\%7Ejl6qk/code/Radar.zip}
	
	\item \textbf{ANOMALOUS} \cite{anomalous_peng2018anomalous} learns the patterns of anomalies by considering the CUR decomposition and residual analysis. A joint learning framework is conducted to select informative attribute to detect anomalies. % \footnote{http://people.virginia.edu/\%7Ejl6qk/code/ANOMALOUS.zip}
	
    \item \textbf{DOMINANT} \cite{dominant_ding2019deep} is a deep learning-based graph anomaly detection method. It leverages a graph autoencoder to reconstruct the adjacency matrix and feature matrix simultaneously to learn the normal patterns of graph. Then, the abnormality of each node is measured by the reconstruction error of node.  % \footnote{https://github.com/kaize0409/GCN\_AnomalyDetection}  
    
    \item \textbf{DGI} \cite{dgi_velickovic2019deep} is a representative self-supervised learning method based on unsupervised contrasting. It learn node representation by maximizing the embedding agreement between each node and the full graph. % \footnote{https://github.com/PetarV-/DGI} \cite{dgi_velickovic2019deep}
    \hl{For this method, we leverage its trained bilinear discriminator with Equation \eqref{eq: contrastive socre} to score node abnormalities.}
    
    \item \hl{\textbf{CoLA} \cite{liuanomaly} is a contrastive self-supervised learning-based anomaly detection method. It captures anomaly patterns by evaluating the agreement between each node and its neighboring subgraph with a GNN-based encoder model.}
\end{itemize}

\begin{table*}[t]
	\small
	\centering
	\caption{The results of ablation study to investigate the effectiveness of each component on $\tool$. Specifically, we use \texttt{SL-GAD-Con}, \texttt{SL-GAD-Gen}, \texttt{SL-GAD-Unweighted}, and \texttt{SL-GAD-Unscaled} to denote \texttt{SL-GAD} without the generative module, contrastive module, anomaly score weighting, and anomaly score scaling, respectively.}
	{
    	\begin{tabular}{p{100 pt}<{\centering}|p{50 pt}<{\centering}p{48 pt}<{\centering}p{45 pt}<{\centering}p{45 pt}<{\centering}p{45 pt}<{\centering}p{45 pt}<{\centering}}%{lcccccc}
    		\toprule
    		Method & BlogCatalog & Flickr & ACM & Cora & CiteSeer & PubMed \\
    		\midrule
    			\tool & \textbf{0.8184} & \textbf{0.7966}  & \textbf{0.8538} & \textbf{0.9130} & \textbf{0.9136} & \textbf{0.9672} \\
    		\midrule
    			\texttt{SL-GAD-Con}\xspace  & 0.7899 & 0.7540  & 0.8308 & 0.8885 & 0.8830 & 0.9482 \\
     			\texttt{SL-GAD-Gen}\xspace & 0.7466 & 0.7442  & 0.7184 & 0.8143 & 0.7841 & 0.7982 \\
     		\midrule
     		    \hl{\texttt{SL-GAD-Unweighted}\xspace}  & \hl{0.8069} & \hl{0.7951} & \hl{0.7972} & \hl{0.9042} & \hl{0.8908} & \hl{0.9419} \\
     		    \hl{\texttt{SL-GAD-Unscaled}\xspace}  & \hl{0.7913} & \hl{0.7812} & \hl{0.8519} & \hl{0.8924} & \hl{0.8670} & \hl{0.9632} \\
    		\bottomrule
    	\end{tabular}
    }
\label{table:ablation}
\end{table*}

\begin{figure*}[t]
\centering
\subfigure[Cora]{
\includegraphics[width=5.5cm]{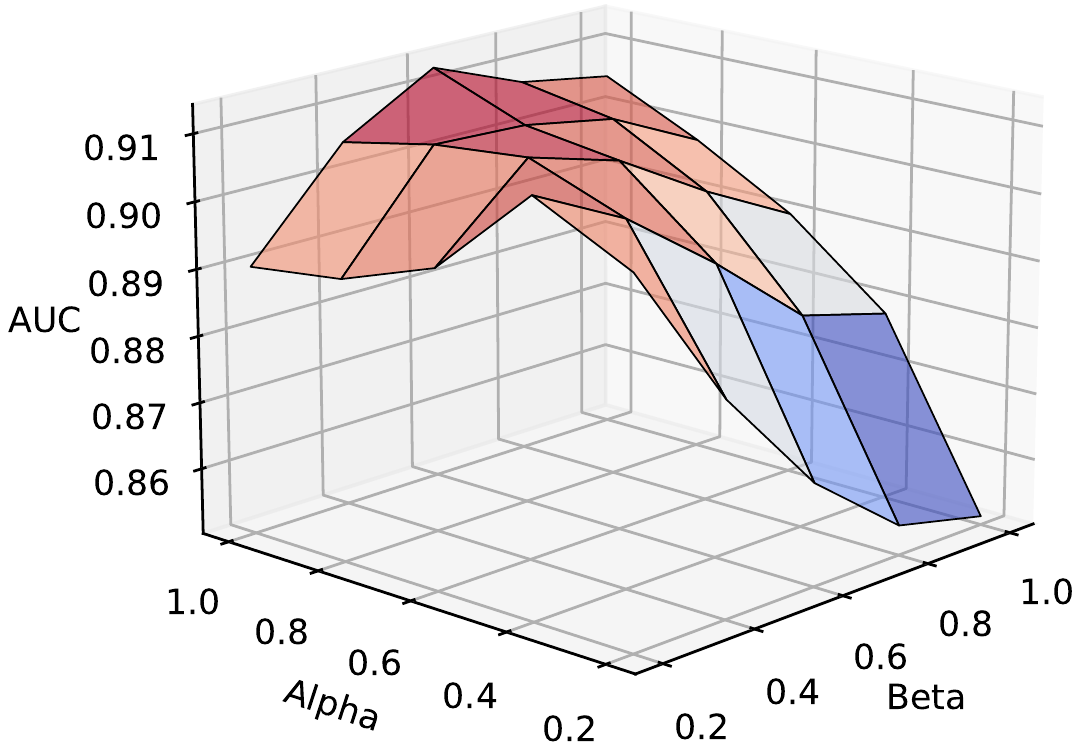}
}
\quad
\subfigure[BlogCatalog]{
\includegraphics[width=5.5cm]{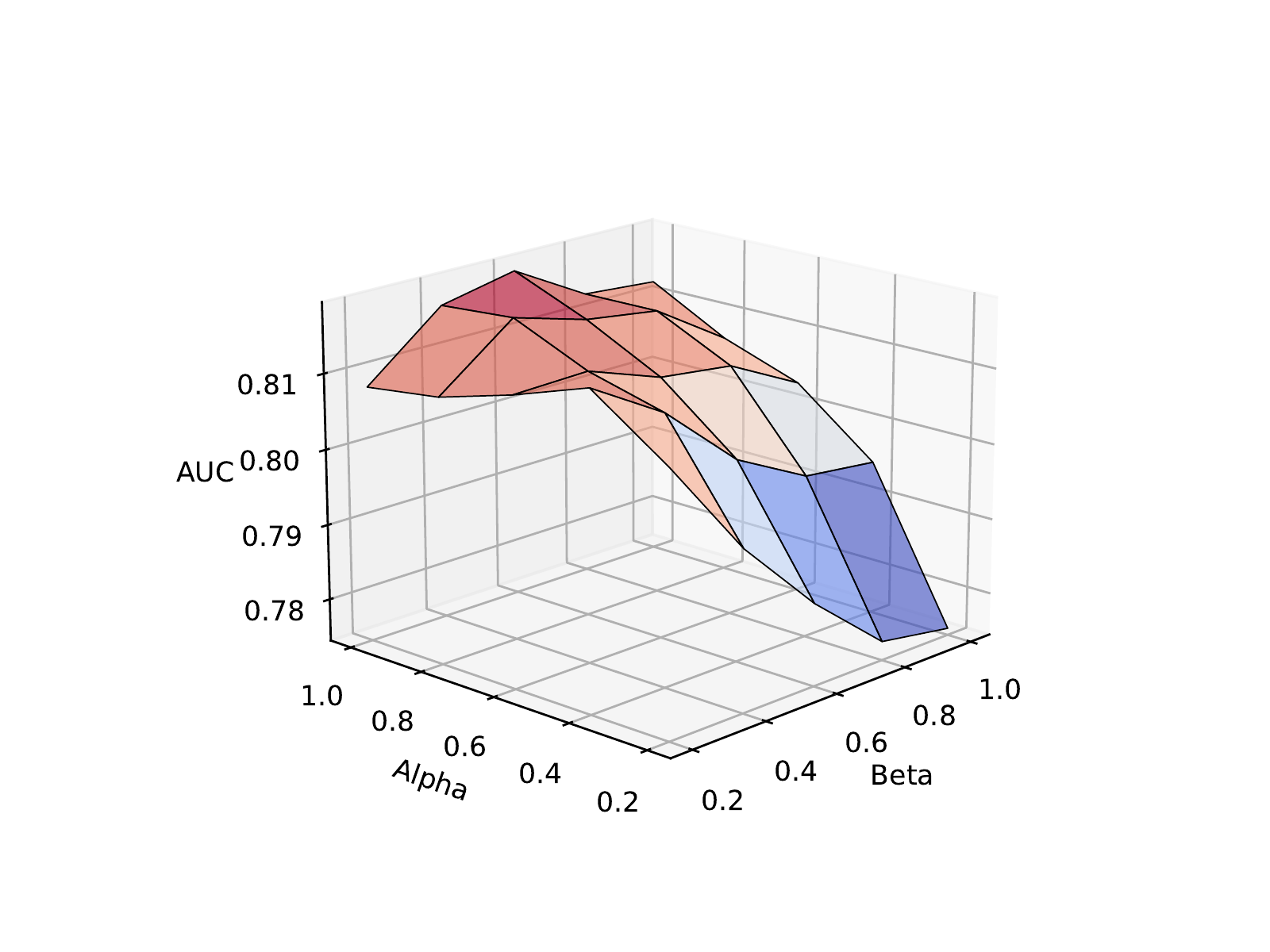}
}
\quad
\subfigure[Flickr]{
\includegraphics[width=5.5cm]{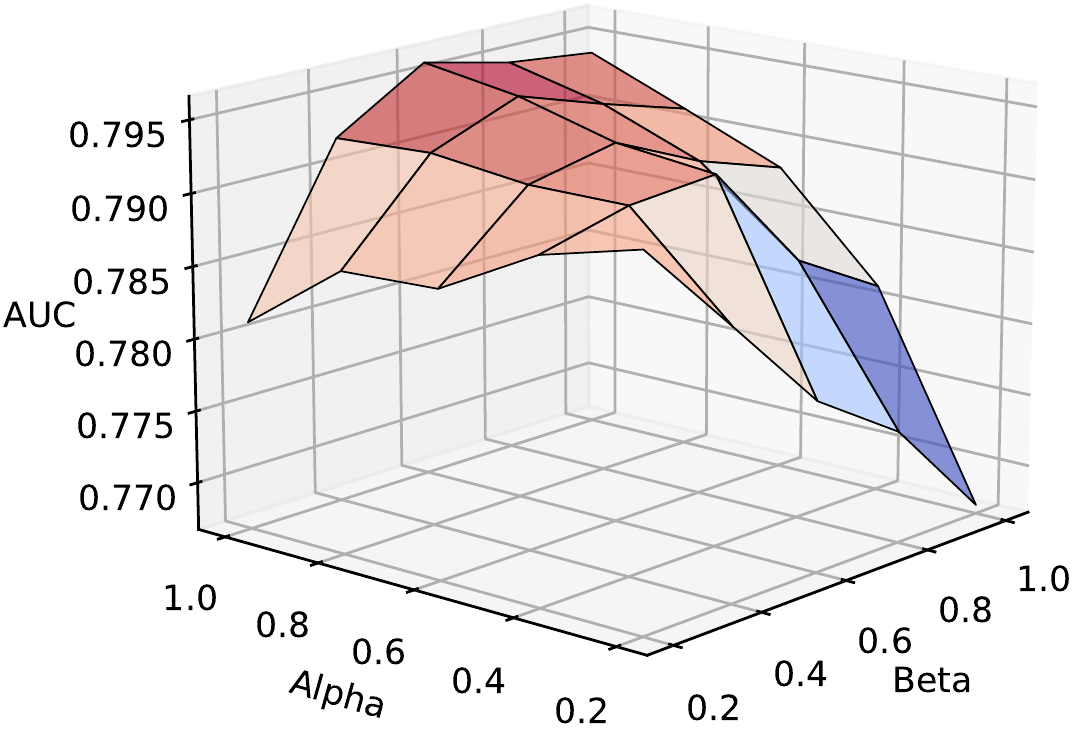}
}
\caption{AUC values of $\tool$ on Cora, BlogCatalog, and Flickr with different weights Alpha and Beta. A warmer color denotes a higher AUC value.}
\label{fig:alpha_beta}
\end{figure*}

\textbf{Metrics.} We utilize ROC-AUC, a widely applied metric for anomaly detection, to quantify the performance of $\tool$ and the competitors. The ROC curve indicates the plot of true positive rate (an anomaly is recognized as an anomaly) against false positive rate (a normal node is recognized as an anomaly). AUC is a value within a range $[0,1]$, which denotes the area under the ROC curve. A larger AUC value indicates a higher detection performance.

\textbf{Parameter Settings.} We set the sampled subgraph size $K$ to be $4$ for efficiency consideration. The GNN encoder is a one-layer GCN model, where hidden dimension $D'$ is $64$. $\alpha$ is fixed to $1$ for all datasets, while $\beta$ is selected from $\{0.2,0.4,0.6,0.8,1\}$. The discrimination modules are trained with Adam optimizer. Cora, Citeseer, Pubmed, and Flickr have learning rates of $0.001$, BlogCatalog has $0.003$, and ACM has $0.0005$. The epoch number for BlogCatalog, Flickr and ACM are $400$, and which for Cora, Citeseer and Pubmed are $100$. The round number $R$ for evaluation is $256$.

\subsection{Comparison with the State-of-the-art Methods} \label{subsec:main_result}

We compare our proposed $\tool$ with the \hl{six} baseline methods in this subsection. The comparison of ROC curves is demonstrated in Figure \ref{fig:roc}, and the results of AUC values are illustrated in Table \ref{table:overall}. According to these results, we summarize our observations as follows:

\begin{itemize} 
	\item The proposed method $\tool$ outperforms all the baseline methods \hl{on all benchmark datasets,} demonstrating its capability to detect anomalies on graph data with high-dimensional node features. The reason behind this is that $\tool$ captures the anomaly patterns by jointly utilizing both generative and contrastive self-supervised learning.
	
	\item The shallow methods (AMEN, Radar, and ANOMALOUS) do not show a competitive anomaly detection performance in our experiments. \hl{This is because these} mechanisms have limited capability to discriminate anomalies from graphs with high-dimensional features and complex structures, which results in relatively poor performance.
	
	\item Compared to other deep methods, $\tool$ shows a stronger detection performance and generalization ability. A possible reason is that these baselines only adopt one learning strategy (e.g., DOMINANT solely uses reconstruction and \hl{CoLA considers contrasting only}), which results in a sub-optimal solution to anomaly detection. In contrast, $\tool$ jointly leverages two self-supervised learning strategies (generative and contrastive), which takes advantage of each other to acquire a higher performance.
	
	\item Our method obtains more considerable performance gains on detecting anomalies in citation networks (ACM, Cora, CiteSeer, and Pubmed) compared to which in social networks (BlogCatalog and Flickr). We observe that these social networks have a more significant degree (mean degree $=32.35$) than citation networks (mean degree $=2.51$), which may lead to an information loss when sampling subgraph views with a fixed size. We leave the efficient view sampling strategy for high-degree graphs as our future work.
\end{itemize}

\begin{figure*}[t]
\centering
\subfigure[Evaluation rounds versus AUC values]{
\includegraphics[width=6cm]{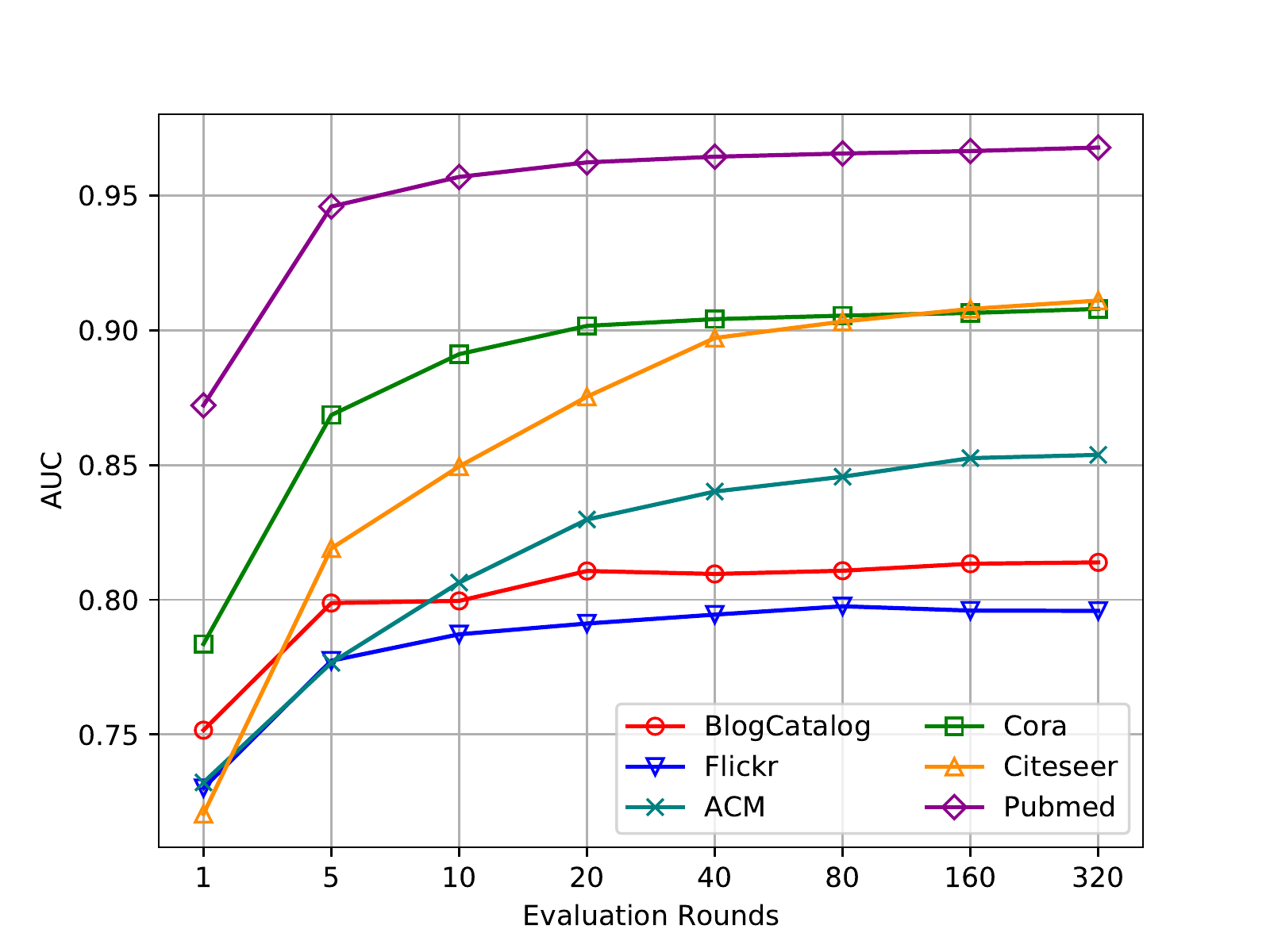}
\label{subfig:r}
}
\subfigure[Subgraph size versus AUC values]{
\includegraphics[width=6cm]{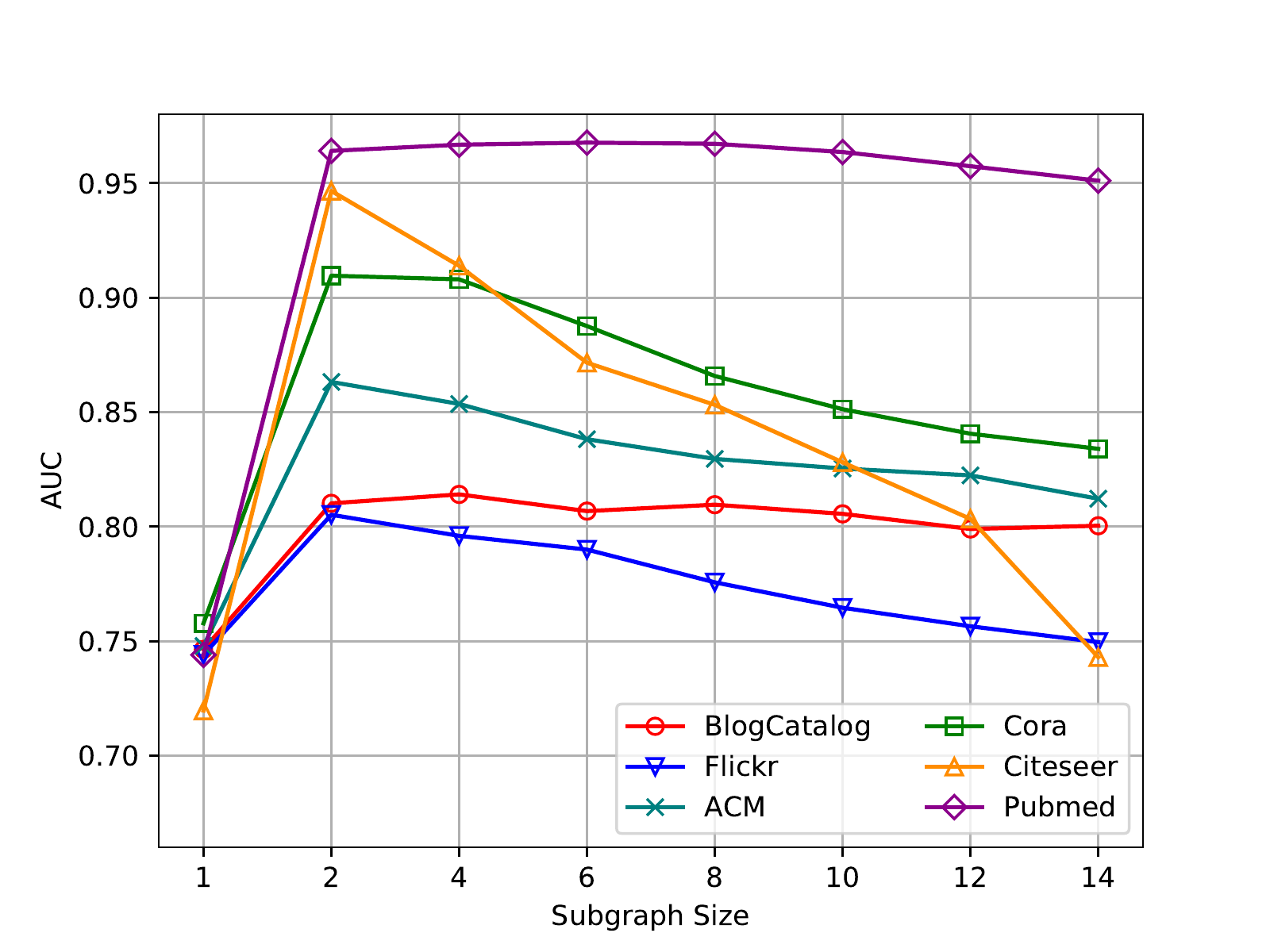}
\label{subfig:k}
}
\subfigure[Hidden dimension versus AUC values]{
\includegraphics[width=6cm]{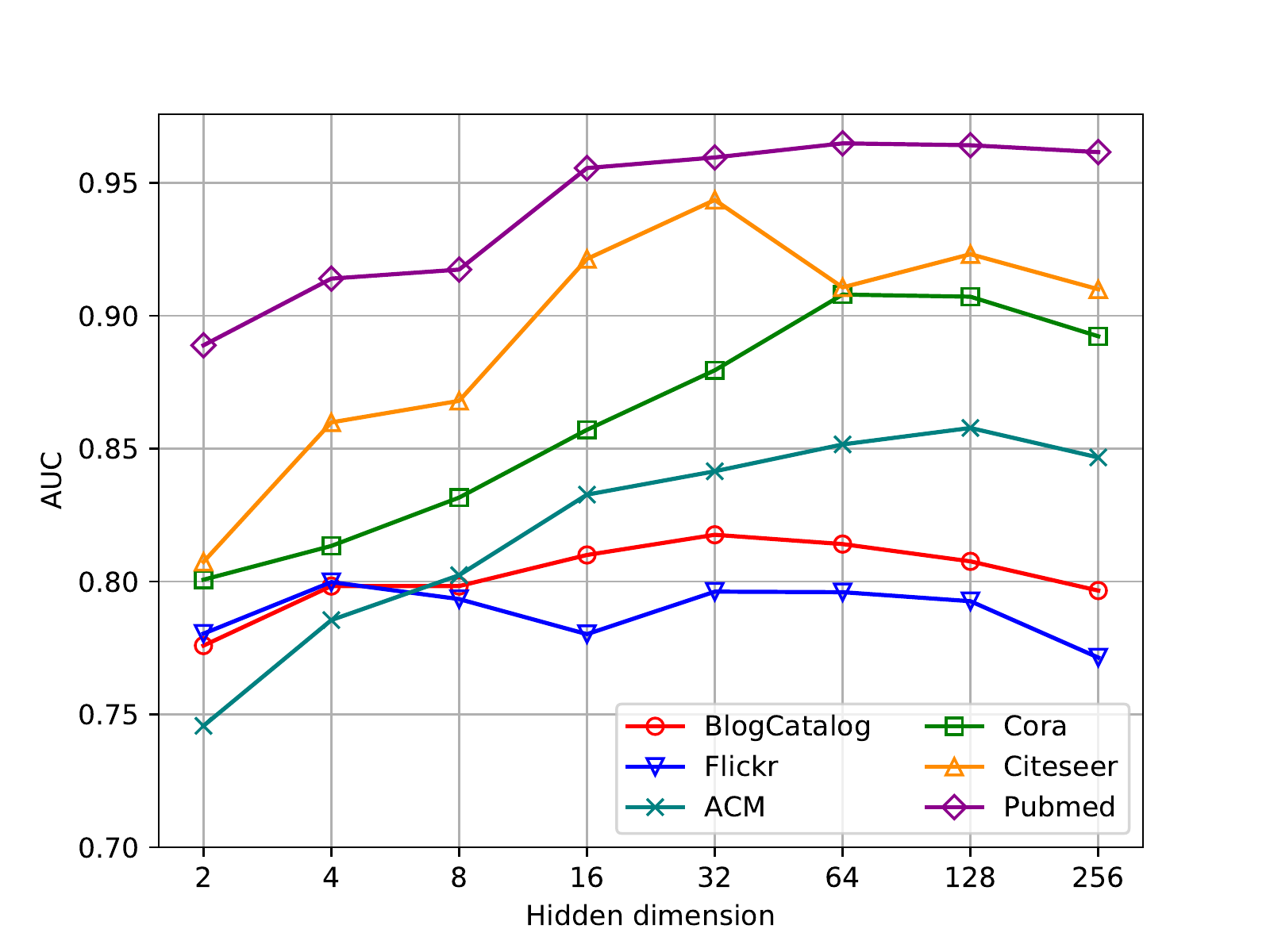}
\label{subfig:d}
}
\subfigure[\hl{Negative ratio versus AUC values}]{
\includegraphics[width=6cm]{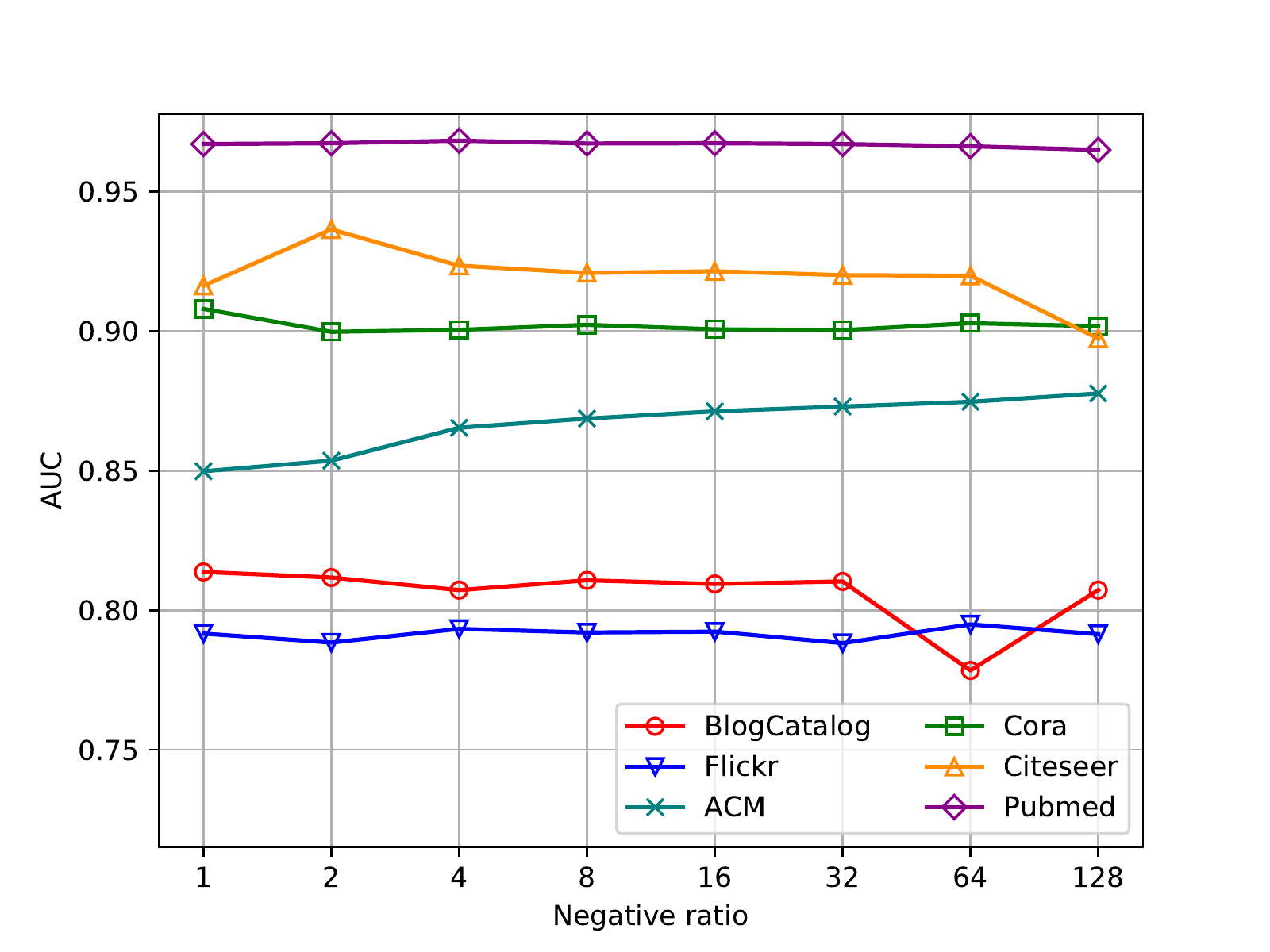}
\label{subfig:n}
}
\caption{
The parameter sensitivity study of $\tool$ on six benchmark datasets. The subfigure (a) shows the performance of our method on different evaluation rounds, where other hyper-parameters are fixed to the default setting as discussed in Section \ref{subsec:exp_setup}. \hl{Similarly, subfigure (b), (c), and (d) studies the impact of subgraph size, hidden dimension, and negative ratio, respectively.}
}
\label{fig:param_sens}
\end{figure*}

\subsection{Effectiveness of Components} \label{subsec:components}

% We investigate the impact of different components of our method in this experiment. We consider two variants of $\tool$ including \texttt{SL-GAD-Con}\xspace only uses contrastive self-supervised learning to identify anomalous nodes but excludes the generative term; \texttt{SL-GAD-Gen}\xspace solely considers feature reconstruction to detect anomalies while the contrastive term is removed from the training loss and anomaly score. We report the results in Table \ref{table:ablation} and conclude the following observations:

We investigate the impact of different components of our method in this experiment. \hl{We first consider two variants of $\tool$, where} \texttt{SL-GAD-Con} only uses contrastive self-supervised learning to identify anomalous nodes but excludes the generative term, and \texttt{SL-GAD-Gen} solely considers feature reconstruction to detect anomalies while the contrastive term is removed from the training loss and anomaly score. \hl{To study the effectiveness of designed anomaly scoring functions, we further construct \texttt{SL-GAD-Unscaled} and \texttt{SL-GAD-Unweighted}, where the first variant removes scalers in Equation \eqref{eq: generative socre} and \eqref{eq: contrastive socre}, and the second variant sums generative and contrastive anomaly scores in an unweighted manner, i.e., $\alpha=\beta=1$ in Equation \eqref{eq: final score}. We report the results in Table \ref{table:ablation} with the following observations:}

\begin{itemize} 

    \item The best performance is achieved by the full $\tool$, which validates the effectiveness of combining contrastive and generative self-supervised learning in a joint learning manner for graph anomaly detection. It also proves that the two self-supervised learning strategies can mutually benefit each other in our method since each of them can recognize exclusive anomaly patterns in graphs.
    
    % \item In all six datasets, \texttt{SL-GAD-Con}\xspace consistently outperforms \texttt{SL-GAD-Gen}\xspace, which demonstrates that contrastive self-supervised learning plays a more critical role in detecting anomalies than generative self-supervised learning. A possible reason is that the agreement between a target node and its neighbouring substructure is highly related to the graph anomalies, which is suggested in \cite{liuanomaly}.
    
    \item In all six datasets, \texttt{SL-GAD-Con}\xspace consistently outperforms \texttt{SL-GAD-Gen}\xspace, which demonstrates that contrastive self-supervised learning plays a more critical role in detecting anomalies than generative self-supervised learning. A possible reason is that the agreement between a target node and its neighbouring substructure is highly related to the graph anomalies, which is suggested in \cite{liuanomaly}. \hl{Compared to the results in Table \ref{table:overall}, both variants can achieve a competitive performance to most baseline methods.}
    
    % \item Compared to the results in Table \ref{table:overall}, both variants can achieve a competitive detection performance to most baseline methods. This indicates that each component of $\tool$ can successfully captures anomalous nodes from real-world graph data via self-supervised learning.
    
    \item \hl{Both anomaly score scaling and weighting facilitate graph anomaly detection on six datasets. When score scaling is disabled, adding generative and contrastive anomaly scores is likely to distort the final measurement because they are formed by different metrics and in different scales. On the other hand, removing score weights also hurts performance. A possible reason is that letting $\alpha=\beta=1$ in Equation \eqref{eq: final score} causes inconsistencies between training and inference because: (1). the calculation of two types of anomaly score relies on trained graph encoder, generative decoder, and contrastive discriminators; (2). $\alpha$ and $\beta$ are not equal to 1 during the training.}

\end{itemize}

% \subsection{Study of Anomaly Scoring} \label{subsec:scoring}

\subsection{Parameters Sensitivity} \label{subsec:parameter}

In this subsection, we carry out a series of experiments to study the effectiveness of various hyper-parameters in $\tool$, including the factors $\alpha$ and $\beta$ to balance generative and contrastive terms, the evaluation rounds $R$, the subgraph size $K$ in graph view sampling, and the dimension of embedding $D'$.

\subsubsection{Balance Factors}

In this experiment, we discuss the impact of the balance factors $\alpha$ and $\beta$ in Eq. (\ref{eq: final score}) and (\ref{eq: obj_func}). We respectively tune the two factors in a range of $\{0.2,0.4,0.6,0.8\}$, and the results are illustrated in Figure \ref{fig:alpha_beta}. Due to limited space, we only show the results on Cora, BlogCatalog and Flickr datasets. According to the results, the AUC values show an upward trend with the increase of $\alpha$, except when $\beta$ is extremely small. Such observation demonstrates that contrastive self-supervised learning is dominant in anomaly detection performance compared with generative self-supervised learning. Furthermore, the selection of $\beta$ highly depends on the specific dataset. For instance, Flickr needs a larger value ($\beta \geq 0.6$) while BlogCatalog prefers a small one ($\beta \leq 0.6$). The results show that it is necessary to find a trade-off between contrastive and generative terms according to the properties of datasets. In practice, we fix $\alpha = 1$ and fine-tune the value of $\beta$ for each dataset.

\subsubsection{Evaluation Rounds}

We investigate the effectiveness of evaluation rounds $R$ in the inference stage of $\tool$. The value of $R$ is selected from $\{1,5,10,20,40,80,160,320\}$. The visualized results are shown in Figure \ref{subfig:r}. It can be observed that the performance is poor when the evaluation rounds are insufficient. With larger $R$, the AUC values rise steadily within a certain range ($R \leq 80$), which indicates that a sufficient number of rounds is essential to prevent the bias caused by randomly sampling. However, when $R$ is large enough ($R \geq 160$), AUC does not have a significant increasing trend even if $R$ is doubled. Consequently, we keep $R$ to be $160$ to acquire a stable performance as well as an acceptable running speed.

\subsubsection{Subgraph Size}

We further explore the sensitivity of subgraph size $K$ in graph view sampling. We run $\tool$ on six datasets for $K = \{1,2,4,6,8,10,12,14\}$ and test the anomaly detection performance. The results are reported in Figure \ref{subfig:k}. 
As we can see from the figure, the performance of $\tool$ increases sharply with the growth of $K$ when $K$ is small and rapidly reaches the peak values. The peak values of AUC appear when $K=2$ for some datasets and $K=4$ for others. After the peak values, the performance drops with the increase of subgraph size. The results show that an appreciated subgraph size is needed to ensure a reliable detection performance. When $K$ is extremely small, the model is hard to acquire sufficient neighboring information to detect anomalies; on the other hand, when $K$ is too large, superfluous information would be included by the subgraph, which hurts the performance.

\subsubsection{Embedding Dimension}

In this experiment, we study the impact of the dimension $D'$ of latent embedding in our GNN-based encoder. The results of varying the values of $D'$ from $2$ to $256$ on six datasets are demonstrated in Figure \ref{subfig:d}. We can observe that in most datasets, the performance \hl{improves} following the increase of hidden dimension when $D' \leq 32$. When $D'$ is larger, there is a peak value of AUC for each dataset, and then the performance drops slightly. The decrease \hl{stems} from the over-fitting problem due to the explosive parameter number. We summarize that $D'$ should be in an appropriate range, e.g., from $32$ to $128$, and finally select $64$ as a general value for all datasets.

\subsubsection{Negative Ratio}

\hl{In this experiment, we change the negative ratio from 1 to 128 to investigate the impact of contrastive negative samples on detection performance. The results are summarized in Figure \ref{subfig:n}, where we define the rate of negatives to positives as the negative ratio, and we have one positive and one negative sample by default in \tool. In general, we observe that increasing the number of negative samples will not significantly affect the detection performance on most of the datasets (e.g., Cora, PubMed, BlogCatalog, and Flickr). A possible reason is that the devised multi-round evaluation mechanism is sufficient for \tool to capture diverse contextual information without relying on a large number of negative samples. On the other hand, unlike other parameter studies, there does not exist a uniform trend regarding the impacts of contrastive negatives on our method, e.g., we found that increasing negatives hurts detection performance on Cora and CiteSeer, but doing so will improve the performance on ACM.}

\section{Conclusion}
% In this paper, we studied the problem of unsupervised graph anomaly detection. We argued that existing approaches did not fully exploit the contextual information of a target node and maximally leverage the available supervision signal in graphs for graph anomaly detection. We proposed a novel self-supervised approach, \tool, for graph anomaly detection. Our approach generates two different subgraph views as contextual information and employs two key components, generative attribute reconstruction and multi-view contrastive learning to fullfill this task. The attribute reconstruction is achieved by a generative learning schema to identify the nodes differing from their neighbors in the attribute space, and the multi-view contrastive learning aims to directly identify the differences between target nodes and their surrounding contexts, thus optimizing these two objectives can capture anomalies in both attribute and structure spaces. Experimental results on six datasets demonstrate the superb performance of the proposed algorithm. 

In this paper, we studied the problem of unsupervised graph anomaly detection. We argued that existing approaches did not fully exploit the contextual information of a target node and \hl{count heavily on available supervision signals} for graph anomaly detection. We proposed a novel self-supervised approach, \tool, for graph anomaly detection. \hl{Our method first generates two different subgraph views of a target node as its contextual information.} \hl{Then, we employ two key components, namely generative attribute reconstruction and multi-view contrastive learning, to fulfill this task.} \hl{The attribute reconstruction leverages a generative learning schema, which identifies} nodes differing from their neighbors in the attribute space\hl{, while the multi-view contrasting tells the difference between nodes and surrounding contexts directly in the hidden} and structure space. Experimental results on six datasets demonstrate the superb performance of the proposed algorithm. 

The proposed method is manually designed for anomaly detection. In the future, we will incorporate  neural architecture search approaches \cite{zhang2020one, zhang2020differentiable, zhang2020overcoming} to automatically design models for anomaly detection.

% or


\ifCLASSOPTIONcaptionsoff
  \newpage
\fi



\bibliographystyle{IEEEtran}
\bibliography{IEEEabrv,content}

\balance

%\vfill


\end{document}